\documentclass[preprint,12pt,authoryear]{elsarticle}
\usepackage{cite}
\usepackage{epsfig}
\usepackage{epstopdf}
\usepackage{epsfig}
\usepackage{amsfonts}
\usepackage{graphicx}
\usepackage{epsfig}
\usepackage{xspace}
\usepackage{color}
\usepackage[abs]{overpic}
\usepackage{makecell}
\usepackage{subcaption}
\usepackage[T1]{fontenc}
\usepackage{mathtools} 
\usepackage{amsmath}
\usepackage{graphicx}
\usepackage{lineno,hyperref}
\usepackage{multirow}
\usepackage[export]{adjustbox}
\modulolinenumbers[1]
\usepackage[margin=2.5cm]{geometry}
\usepackage{verbatim}

\usepackage{soul,microtype}

\usepackage{xcolor}

\journal{Remote Sensing of Environment}

\begin{document}

\begin{frontmatter}
\title{Very High-Resolution Forest Mapping with \mbox{TanDEM-X} InSAR Data and Self-Supervised Learning}

\author[dlr]{Jos\'e-Luis Bueso-Bello}
\author[dlr,ensta]{Benjamin Chauvel}
\author[dlr]{Daniel Carcereri}
\author[dlr]{Philipp Posovszky}
\author[dlr,houston]{Pietro Milillo}
\author[houston2]{Jennifer Ruiz}
\author[houston]{Juan-Carlos Fern\'andez-Diaz}
\author[dlr]{Carolina Gonz\'alez}
\author[dlr]{Michele Martone}
\author[dlr]{Ronny Hänsch}
\author[dlr]{Paola Rizzoli}

\affiliation[dlr]{organization={German Aerospace Center (DLR), Microwaves and Radar Institute},
            city={Weßling},
            country={Germany}}

\affiliation[ensta]{organization={ENSTA Bretagne},
            city={Brest},
            country={France}}
            
\affiliation[houston]{organization={Department of Civil and Environmental Engineering, University of Houston},
            city={Houston, TX},
            country={USA}}

\affiliation[houston2]{organization={Department of Biology and Biochemistry, University of Houston},
            city={Houston, TX},
            country={USA}}

\begin{abstract}
Deep learning models have shown encouraging capabilities for mapping accurately forests at medium
resolution with \mbox{TanDEM-X} interferometric SAR data. Such models, as most of current state-of-
the-art deep learning techniques in remote sensing, are trained in a fully-supervised way,
which requires a large amount of labeled data for training and validation. In this work, our aim is to exploit the high-resolution capabilities of the \mbox{TanDEM-X} mission to map
forests at 6 m. The goal
is to overcome the intrinsic limitations posed by mid-resolution products, which affect, e.g.,
the detection of narrow roads within vegetated areas and the precise delineation of forested regions contours. To cope with
the lack of extended reliable reference datasets at such a high resolution, we investigate self-supervised learning techniques
for extracting highly informative representations from the input features, followed by a supervised training step with a significantly
smaller number of reliable labels. A 1 m resolution forest/non-forest reference
map over Pennsylvania, USA, allows for
comparing different training approaches for the development of an effective forest mapping
framework with limited labeled samples. We select the best-performing
approach over this test region and apply it in a real-case forest mapping scenario
over the Amazon rainforest, where only very few labeled data at high resolution are available.
In this challenging scenario, the proposed self-supervised framework significantly enhances
the classification accuracy with respect to fully-supervised methods, trained using the same
amount of labeled data, representing an extremely promising starting point for large-scale,
very high-resolution forest mapping with \mbox{TanDEM-X} data.
\end{abstract}


\end{frontmatter}



\section{Introduction}
\label{sec:introduction}

Forests are of paramount importance for the Earth's ecosystem, since they play a key role in reducing the
concentration of carbon dioxide in the atmosphere and in controlling climate change \citep{unfccc_2020, Schepaschenko2021}.
Human activities, such as selective logging, illegal deforestation, and natural hazards, 
impact forest health and can lead to forest degradation and loss.
Therefore, a reliable assessment and monitoring of
forested areas at large- and global-scale is of critical importance for assessing forest resources and properly  informing decision-making stakeholders \citep{fao_global_2020}. 

In this context, satellite remote sensing represents a powerful tool for mapping forests and their properties on a large and global scale \citep{10.1093/forestry/cpad024}.
In recent years, the continuous availability of remote sensing data and their increasing resolutions and revisit times
have allowed for the generation of global forest/non-forest maps, mainly derived from multi-spectral optical data.
In \citep{Hansen2013} a complete world forest coverage map at 30~m resolution is derived from Landsat multi-spectral data, including forests changes detected between 2000 and 2023. 
More recently, global land cover maps, mainly based on Sentinel-2 multi-spectral imagery at 10~m resolution and including a forest class, have been released, such as the
Finer Resolution Observation and Monitoring of Global Land Cover (FROM-GLC) map from 2017 \citep{gong2019FROM-GLC}, the global ESA WorldCover for 2020 and 2021 \citep{esa_worldcover_2020, esa_worldcover_2021} and the large-scale maps from the ESA Climate Change Initiative (CCI) High Resolution Land Cover (HRLC) project \citep{esa_hrlc_cci_2024}.
However, approaches solely based on optical data may suffer from the presence of clouds, with an estimated 50\% of the Earth's surface being hidden by clouds at any given moment \citep{gawlikowski_clouds_2022}. In this context, Synthetic Aperture Radar (SAR) systems represent an attractive solution due to their capability of acquiring data almost independently of weather and daylight conditions.
The first global forest coverage map based on SAR images was generated from ALOS-PALSAR satellite data at L-band, based on cross-polarization backscatter images and provided at a resolution of 25~m \citep{shimada2014forest}.
More recent investigations relying on the ESA Sentinel-1 data at C-band have also demonstrated the great potential of dual-polarization acquisitions to monitor forests \citep{hansen_s1_fnf_2020, ricardo_cnn_s1_2022}. 
In addition, the enhanced capabilities of interferometric SAR (InSAR) systems to monitor vegetated areas, and especially the added value of the interferometric coherence, have been demonstrated in \citep{SCHLUND201416, martone2018FNF}.

In the context of global InSAR datasets, the \mbox{TanDEM-X} (TerraSAR-X add-on for Digital Elevation Measurement) mission represents the first spaceborne InSAR mission acquiring bistatic images at X-band over the complete Earth's landmasses. The two twin satellites TerraSAR-X and \mbox{TanDEM-X} have been flying in close orbit formation since 2010, constituting a single-pass interferometer with variable baselines and acquisition geometries \citep{krieger2007tandem, zink_tdm_2021}. 
The main goal of the mission is the generation of a global Digital Elevation Model (DEM) at a spatial resolution of 12~m, 
which was completed successfully in 2016 \citep{rizzoli2016DEMperformance, gonzalez2018}.
Beside the nominal DEM product, for each \mbox{TanDEM-X} bistatic acquisition, additional bypass products, such as the calibrated backscatter, the interferometric phase and the interferometric coherence, are available as well. More concretely, the volume correlation factor derived from the interferometric coherence \citep{martone2016volume, rizzoli_volume_2022}, was the main input feature for the generation of the global \mbox{TanDEM-X} Forest/Non-Forest (FNF) map at 50~m resolution, based on a fuzzy clustering machine learning algorithm \citep{martone2018FNF}. 
Additionally, local maps at national scale were generated at a finer resolution 
(12~m) using an enhanced version of the forest classification approach, aimed at preserving both global classification accuracy and local precision thanks to the introduction of nonlocal filtering for the estimation and denoising of the interferometric coherence \citep{martone2018FNFhighres}.

In the last few years, deep learning (DL) approaches have started to significantly impact spaceborne SAR applications \citep{zhu2017DLinSAR, ma2019DLRS}. 
In particular, Convolutional Neural Networks (CNNs) trained in a fully-supervised way have shown great potential for the extraction of informative patterns from SAR images in various application fields, such as semantic segmentation for land cover classification \citep{zhu2021DLSAR,ricardo_cnn_s1_2022}, forest parameter retrieval \citep{Carcereri2023,CARCERERI2024114270}, or SAR signal processing and image enhancement \citep{phinet, Sica2022, Pulella2024}.
Regarding the specific case of forest mapping using \mbox{TanDEM-X} and DL, preliminary works are presented in \citep{Mazza2019} and \citep{buesobello_fnf_2022}, where the proposed models, based on a U-Net architecture, are trained in a fully-supervised manner using mid-resolution input data varying from 12~m up to 50~m for local and large-scale products, respectively. 

When moving to finer resolutions or specific application domains, the lack of reliable reference datasets has boosted the investigation of self-supervised learning (SSL) approaches \citep{ssl_survey}.
However, to the best of our knowledge, no studies have been published in the literature yet that make use of SSL for land cover-related applications using spaceborne InSAR data. 

The objective of this study is to investigate the potential of SSL methods for forest mapping with \mbox{TanDEM-X} InSAR data processed down to 6~m resolution (independent pixel spacing) and to benchmark them with respect to a traditional fully-supervised learning approach.
We first select the entire state of Pennsylvania, USA, as study site for which a reference forest map from 2010 at 1~m resolution is available. The large extension of such a reference map allows for the set up of a baseline DL model derived through fully-supervised learning (FSL), which, in the presence of a sufficiently large amount of labeled data, represents the best-case scenario. As in previous works \citep{Mazza2019, buesobello_fnf_2022}, we rely on a U-Net model for the fully-supervised analysis. Afterwards, we investigate different SSL approaches and assess their suitability as pretext task to forest mapping, using the same input features as for the fully-supervised method. In particular, the goal of the SSL approach is to train a DL model 
that maps an image $x_1$ to a representation of visual contents $\hat{x}_1$ without the necessity for annotated data. As the starting DL architecture, we consider a classic convolutional autoencoder (CAE) and we investigate two different pretext tasks which aim at reconstructing the input features via a standard identity reconstruction and a masked CAE, denoted here as \textit{identity} and \textit{inpainting} tasks, respectively. While both approaches aim at reconstructing the original input feature maps, the masked CAE has to tackle the additional challenge that part of the input is artificially occluded.
In a second step, a supervised learning phase is necessary to perform the downstream task of forest mapping.
Regarding the DL model architecture, we rely on a U-Net as in the case of the FSL baseline model but, in this case, the encoder part is initialized with 
the weights from the encoder part of the CAE trained in an SSL manner. 
In this study 
we distinguish between two different uses of the U-Net, depending on the initialization of the encoder weights:
When a random initialization is considered, the U-Net is trained in a classic fully-supervised learning (FSL) manner. 
Differently, when we transfer knowledge from the CAE trained using an SSL pretext task, we refer to the use of the U-Net as downstream task (DST).
To assess the impact of the SSL pre-training on the downstream task, as well as to find a compromise between the final performance and
the amount of required reference data to reach it,
we investigate different scenarios based on: 
a) the type of pretext task used in the SSL part (identity or inpainting); 
b) the type of training after transferring the weights from the SSL to the supervised DST part (partial or full trainability of the U-Net after initialization with the SSL weights);
c) the usage of a reduced amount of labeled data in the supervised DST
part selected from the ones used for the baseline FSL scenario.
Finally, we apply and validate the best-performing SSL+DST approach to a real case scenario over the Amazon rainforest, where the use of pure FSL is jeopardized by the lack of extended reference data. 

This paper is organized as follows: Section~\ref{sec:materials} presents the 
main characteristics of the \mbox{TanDEM-X} interferometric dataset and the LiDAR and optical reference data used for training, testing and comparison.
Section~\ref{sec:methods}
details the DL architectures proposed in our work, the different SSL and FSL methodologies for training, validation and testing, and the performance metrics adopted for the accuracy assessment.
The classification results for the different DL scenarios on single \mbox{TanDEM-X} acquisitions over the Pennsylvania study site are presented in Section~\ref{sec:results}. As real-case application scenario of our methodology, this section also includes the results obtained over the Amazon rainforest, together with an comparison to a 10~m large-scale forest map based on optical data.
The discussion of the conducted deep-learning experiments and the achieved results is detailed in Section~\ref{sec:discussion}. Finally, in Section~\ref{sec:conclusions}
the conclusions and outlook to future work are drawn.
\section{Materials}
\label{sec:materials}

\subsection{TanDEM-X InSAR dataset}
\label{ssec:tdx_dataset}

\begin{figure}[t]
    \centering
    \includegraphics[width=.95\textwidth]{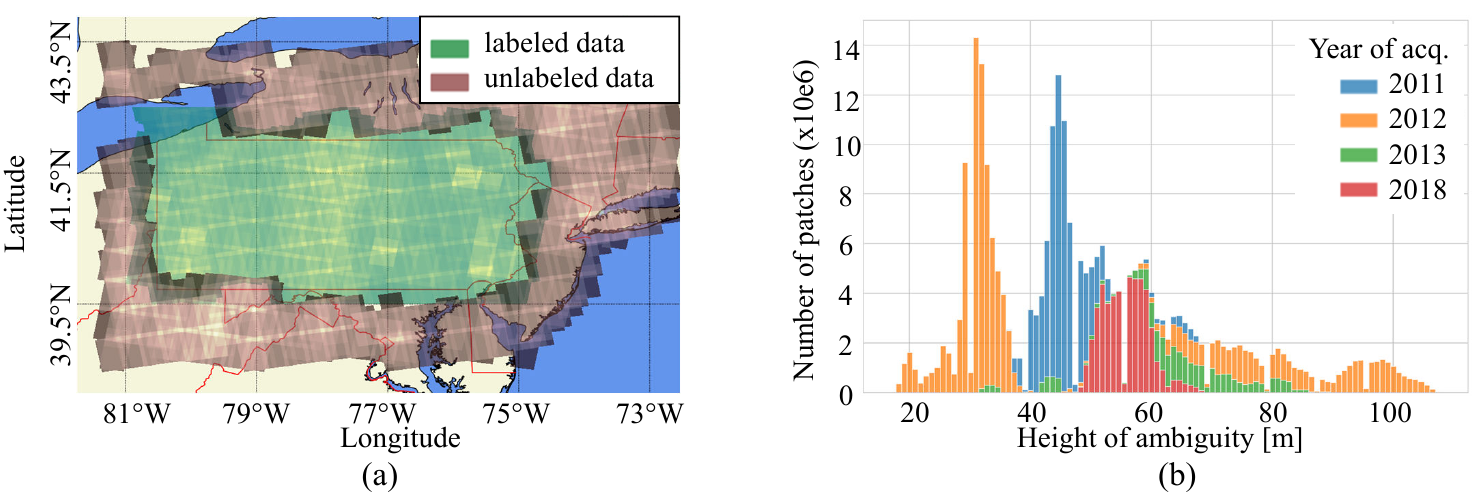}
    \caption{TanDEM-X acquisitions used in this study over the the Pennsylvania state, USA. (a) Ground coverage and (b) Total $h_{\mathrm{amb}}$ distribution. The individual colors show the relative patch contribution of the different years to the total.}
    \label{fig:data}
\end{figure}

\mbox{TanDEM-X} is the first operational spaceborne InSAR mission comprising two different spacecrafts, namely the twin satellites TerraSAR-X and \mbox{TanDEM-X}. It has been globally acquiring 
InSAR images in a bistatic configuration since the end of 2010 \citep{krieger2007tandem, zink_tdm_2021}. 
In this work, we rely on specific collections of \mbox{TanDEM-X} InSAR images acquired over two different regions:

\begin{itemize}
    \item \textit{The Pennsylvania study site:} We consider more than 1500 \mbox{TanDEM-X} bistatic images acquired over temperate forests, including 450 data takes over the state of Pennsylvania, USA.
\mbox{Figure~\ref{fig:data}} shows the ground coverage (a) and the distribution of the corresponding height of ambiguity $h_{\textrm{amb}}$ (b). This parameter is linked to the InSAR acquisition geometry, as explained later on in \eqref{eq: HoAeqn}. 
TanDEM-X acquisitions overlaying the reference data area 
are shown in green in \mbox{Figure~\ref{fig:data}(a)}. 
These were acquired between 2011 and 2012, to reduce the temporal separation with respect to the available reference data, detailed in Section~\ref{ssec:ref_map_penn}. 
Further \mbox{TanDEM-X} acquisitions, shown in brown, are considered to increase the size of the SSL dataset in our study. They are acquired over temperate forests with similar characteristics to the ones in the reference data area
and are representative of the global variability of $h_{amb}$ values (\mbox{Figure~\ref{fig:data}(b)}).

    \item \textit{The Amazon rainforest application scenario:} We divide the \mbox{TanDEM-X} data over the Amazonas in three different subsets as follows.
    \begin{itemize}
            \item \textit{TanDEM-X images for SSL:}   
    670 \mbox{TanDEM-X} scenes mainly acquired over the
    Acre, Rondônia, Mato Grosso, and Pará states in Brazil. This area is also known as "arc of deforestation" \citep{amazonas_arc_defo2013}. To ensure a balanced dataset between forest and non-forest areas, we consider only heterogeneous \mbox{TanDEM-X} scenes with at least 25\% coverage for each of the two land cover classes.
    
\item \textit{TanDEM-X images for the downstream task:}
35 \mbox{TanDEM-X} images for the supervised DST step of the proposed framework. 

\item \textit{TanDEM-X images for maps intercomparison:}
500 \mbox{TanDEM-X} images acquired between 2019 and 2020 with different geometries over the south-east part of the Amazon rainforest.
    
\end{itemize}
    \end{itemize}

The \mbox{TanDEM-X} co-registered single-complex (CoSSC) products, which
were previously focused and co-registered with the operational integrated \mbox{TanDEM-X} processor (ITP) \citep{fritz_itp_2012},
are the input to our processing chain.  
For each \mbox{TanDEM-X} bistatic acquisition, besides the nominal InSAR DEM, 
additional SAR and InSAR quantities are available.
As in previous works \citep{buesobello_fnf_2022}, we rely on the absolutely calibrated backscatter, the interferometric coherence, and the volume correlation factor as the main input features for our investigations.

The absolutely calibrated backscatter $\beta_0$ corresponds to the radar brightness recorded by the transmitting satellite.

For the estimation of the InSAR coherence at 6~m, we rely on the application of $\Phi$-Net, a state-of-the-art residual DL model for InSAR parameter estimation and denoising \citep{phinet}.
The bistatic interferometric coherence gives information about the amount of noise in the interferogram. As described in \citep{zebker1992decorrelation}, several error sources may contribute to coherence loss,
which, assuming statistical independence, can be factorized as follows \citep{martone2012Coh_ISPRS}:
\begin{equation}
\gamma_\mathrm{Tot} = \gamma_\mathrm{SNR}\cdot\gamma_\mathrm{Quant}\cdot\gamma_\mathrm{Amb}\cdot\gamma_\mathrm{Rg}\cdot\gamma_\mathrm{Az}\cdot\gamma_\mathrm{Vol}\cdot\gamma_\mathrm{Temp}.
\label{eq: coherence_tot}
\end{equation}
\noindent where the terms on the right-hand side of the equation identify different decorrelation factors caused by limited SNR ($\gamma_\mathrm{SNR}$), quantization errors ($\gamma_\mathrm{Quant}$), range and azimuth ambiguities ($\gamma_\mathrm{Amb}$), baseline decorrelation ($\gamma_\mathrm{Rg}$), relative shift of the Doppler spectra ($\gamma_\mathrm{Az}$), volumetric scattering effects ($\gamma_\mathrm{Vol}$) and temporal changes ($\gamma_\mathrm{Temp}$).
As \mbox{TanDEM-X} operates as a single-pass radar interferometer, it is not affected by temporal decorrelation, i.e. $\gamma_\mathrm{Temp} = 1$. Differently, $\gamma_\mathrm{Vol}$ is not negligible 
in the presence of volumetric targets, such as vegetation or snow/ice-covered areas \citep{martone2016volume}. For its computation, we apply the procedure detailed in \citep{rizzoli_volume_2022}.

To represent the large variety of acquisition geometries possible in the \mbox{TanDEM-X} mission, we rely on the 
height of ambiguity $h_{\mathrm{amb}}$
and the local incidence angle $\theta_\mathrm{i}$.

The $h_{\mathrm{amb}}$ represents the topographic height difference corresponding to a complete $2\pi$ cycle of the interferometric phase \citep{martone2016volume}. 
For the bistatic case, it is defined as:
\begin{equation}
h_{\mathrm{amb}} = \frac{\lambda \cdot {r} \cdot \mathrm{sin}(\eta_i)}{B_\perp},
\label{eq: HoAeqn}
\end{equation}
with $\lambda$ being the radar wavelength, $r$ the slant range, $\eta_i$ the acquisition incidence angle and $B_\perp$ the baseline perpendicular to the line of sight. 
The majority of nominal \mbox{TanDEM-X} acquisitions is characterized by $h_{\textrm{amb}}$ values ranging from 20~m to 120~m.
The local radar waves incidence angle represent the incident angle normal to the surface direction and it can be derived by knowing the underlying topography \citep{rizzoli_volume_2022}. For its computation, we rely on the use of the 30~m edited \mbox{TanDEM-X} DEM \citep{gonzalez_editing_2020}.

\subsection{Reference forest map over Pennsylvania}
\label{ssec:ref_map_penn}
For this study, we consider as high-resolution reference data a forest/non-forest map over the Pennsylvania state, USA - 
the result of a joint collaboration between the University of Maryland and the University of Vermont \citep{lidar_data}.
LiDAR and optical data acquired up to 2010 are combined to generate a binary reference map at a ground resolution of 1~m, classifying forests as vegetation higher than 2~m. 
This map was originally used for the validation of the Landsat forest map and an accuracy of about 98\% is reported. 
To match the resolution used in our study, we scale the
original resolution down to 6~m.

\subsection{LiDAR reference data over the Amazon rainforest}
\label{ssec:ref_data_ama}

For our application scenario over the Brazilian Amazon, we consider as high-resolution reference data 
different forest/non-forest patches over the states of Pará and Mato Grosso, Brazil, derived from LiDAR airborne data.
The original point-cloud data were acquired between 2012 and 2018 as small footprint LiDAR surveys over selected forest research sites across the Amazon rainforest in Brazil as part of the Sustainable Landscapes Brazil Project and made accessible through the NASA/ORNL Distributed Active Archive Center (DAAC) \citep{patches_amazonas}. 
The original point cloud files in LAS format were downloaded and refined for this work by the National Center for Airborne Laser Mapping (NCALM) at the University of Houston, USA. The multiple LAS files for a given area of interest were retiled into TerraScan projects. Afterwards, the ground classification was densified and the distance for each return from the ground model was computed. This distance was used to classify the returns into close-to-ground ($\pm25$~cm) low, medium and high vegetation strata. The distance from ground was also used to produce a Canopy Height Model (CHM) at 1 m raster grid spacing based on the largest distance for each cell.
The raster were produced using the original point cloud UTM projection, but for our investigations, we reprojected these rasters to the EPSG:4326 coordinate reference system and downsampled them to the desired 6~m resolution in this study.
Finally, to generate a reference binary forest/non-forest map, we empirically set a threshold at 4~m over the CHMs as a good trade-off between keeping high-resolution details and respecting the uniformity of dense forested areas and in accordance to \citep{isForest_Chazdon_2016}.

\subsection{ESA High Resolution Land Cover map}
\label{ssec:esa_hrlc_cci}

For intercomparison of the obtained forest/non-forest maps from \mbox{TanDEM-X} images, we 
consider the map produced
by the ESA Climate Change Initiative (CCI) High Resolution Land Cover (HRLC) project, covering the geographic range: 13.5$^\circ$S – 7.5$^\circ$S and 51.5$^\circ$W – 62.1$^\circ$W.
The HRLC map is mainly based on Sentinel-2 images acquired during 2019 and is provided at a 10~m spatial resolution \citep{esa_hrlc_cci_2024}.
It contains 15 different land cover classes, including different tree cover types, shrubs, croplands, grasslands, bare land, built-up areas and open water. We classify all four tree cover classes as forest, including evergreen and deciduous trees, as well as broadleaf and needleleaf ones. We set the pixels belonging to the other land cover classes as non-forest.

\section{Methods}
\label{sec:methods}

In this section we present the details of the developed approach for forest mapping at high-resolution with \mbox{TanDEM-X}.
We first describe the proposed DL models architectures, the complete set of input variables and the different training strategies utilized for 
the convolutional autoencoder (CAE) and the U-Net, respectively.
Finally, we introduce the performance metrics employed to evaluate the results.

\subsection{Proposed deep learning framework}
\label{ssec:cnn_architectures}

We use two different CNN models in our study: a convolutional autoencoder (CAE) in the SSL part and a U-Net for the task of forest mapping by means of either fully-supervised learning or training after domain transfer from the encoder part of the CAE (\mbox{Figure~\ref{fig:dl_architectures}}).

\subsubsection{CAE and U-Net architectures}
\label{sssec:ssl_model}

\begin{figure}
    \centering
    \includegraphics[width=.8\textwidth]{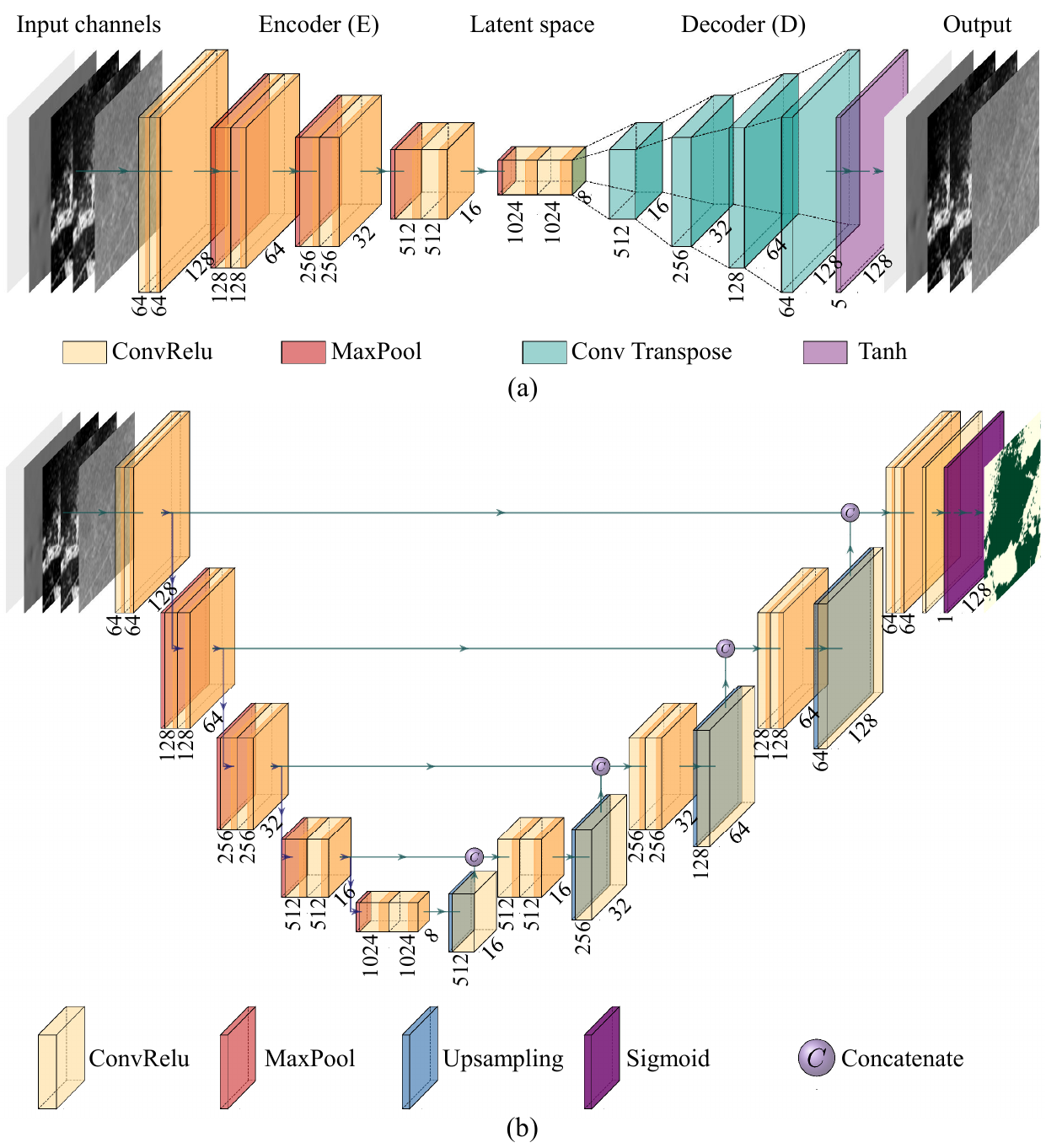}
    \caption{CNN architectures used in the study: (a) Convolutional autoencoder (CAE) and (b) U-Net. The structure of the encoding part is common to both CNNs.}
    \label{fig:dl_architectures}
\end{figure}

A CAE, as depicted in \mbox{Figure~\ref{fig:dl_architectures}}(a), is a type of feedforward neural network composed of two parts: a contracting path (also known as the encoder) and an expanding path (also known as the decoder). The contracting path is responsible for 
the extraction of high-level representations of the input, 
while the expanding path is responsible for 
recovering the spatial resolution and producing the desired output.

In our encoder, each convolutional level consists of two consecutive convolutions with $3\times3$ pixel kernels, each followed by batch normalization and a Rectified Linear Unit (ReLU) activation function.
The last layer of each convolutional block performs a $2\times2$ max-pooling operation to reduce the spatial resolution. For the initial convolutional layer we apply 64 filters, which are doubled at each successive level (\mbox{Figure~\ref{fig:dl_architectures}}(a)).
The decoder of the CAE 
mirrors the encoder and consists of 
2D transposed convolution operators 
with $3\times3$ pixel kernels to upsample the features extracted from the encoder. 
A batch normalization and a ReLU activation function follows after each transposed convolution.
The final layer has a hyperbolic tangent activation function (Tanh) to achieve 
a faster convergence than using a sigmoid activation function. 

Regarding the U-Net architecture \citep{ronneberger2015unet}, we consider the same encoding part of the CAE and we add skip connections to concatenate the feature maps at the different layers of the encoder to the corresponding level in the decoder (\mbox{Figure~\ref{fig:dl_architectures}}(b)). This choice is driven by the fact that this allows us to directly transfer the encoder weights of the CAE, trained during the SSL task, to the encoder of the U-Net.
For the decoder of the U-Net, we perform an upsampling on every convolutional level by applying a 
$2\times2$ transposed convolution, followed by a batch normalization and a ReLU activation function. 
We concatenate the outputs of each convolutional level with the corresponding encoder features and go through two successive convolutions, each one followed by a batch normalization and a ReLU transformation.
For the last convolutional layer we apply a $1\times1$ convolution with a
sigmoid activation function to map 64 input features into the probability of each pixel to be forest. 

Finally, for all DL models presented in our study, we consider as input a stack of 128 $\times$ 128 pixels patches obtained from the \mbox{TanDEM-X} InSAR dataset and representing the following quantities:
\begin{itemize}
    \item \textit{SAR features}: absolutely calibrated backscatter $\beta_0$, 
    \item \textit{InSAR features}: bistatic coherence $\gamma_{\mathrm{Tot}}$ and volume correlation factor $\gamma_{\mathrm{Vol}}$,
    \item \textit{Acquisition geometry descriptors}: local incidence angle $\theta_{\mathrm{i}}$ and height of ambiguity $h_{\mathrm{amb}}$.
\end{itemize}
\noindent 
We filter out all pixels affected by geometric distortions, such as shadow and layover, and the different input channels are normalized internally by the Pytorch transformer \citep{torch}.

\subsection{SSL pretext tasks using CAE}

We investigate the use of two different SSL pretext tasks: \emph{identity} and \emph{inpainting}. Their description is provided in the following subsections and, for the sake of brevity, in the remaining of the paper they are identified as \textbf{SSL-Id} and \textbf{SSL-In}, respectively.

\subsubsection{\textbf{SSL-Id}: Identity task}

The \textit{identity task} aims at using the CAE to faithfully reconstruct the input channels $\textbf{x}$ by minimizing the distance between the network's input itself and the output $F(\textbf{x})$. 
As loss function, we utilize a two-term function defined as:
\begin{equation}
        \mathcal{L}_{\mathrm{identity}} = \text{L}_1 + \text{L}_2,
\end{equation}
where 
L$_1$ identifies the L1-norm of the prediction error: 
\begin{equation}
\label{eq:l1}
        \text{L}1(\textbf{x}) = \lVert \textbf{x}-F(\textbf{x})\rVert_1,
\end{equation}
and L$_2$ represents the L2-norm of the prediction error:
\begin{equation}
\label{eq:l2}
        \text{L}2(\textbf{x}) = \lVert \textbf{x}-F(\textbf{x})\rVert_2.
    \end{equation}
 
\noindent 

Their combination effectively weights the impact of outliers on the solution.

\subsubsection{\textbf{SSL-In}: Inpainting task}
The \emph{Inpainting} task can be performed by using \emph{masked autoencoders}, also known as \textit{context encoders} \citep{pathak2016context_encoders_inpainting}.
In this case, part of the input features set ${\textbf{x}}$ is masked out prior to be given to the model, which tries to predict the missing parts 
by learning from the spatial context around the masked area
\citep{singh2018ssl_semantic_segmentation_overhead_imagery}. 
In our study, we use a binary mask $M$ of size 128~pixels~$\times$~128~pixels, in which 
we mask an area of size 64~pixels~$\times$~64~pixels randomly located inside the input patches.
The masked area in $M$ has a pixel value of 0, 1 elsewhere.
Then, the input of the SSL model $\hat{\textbf{x}}$ is estimated as:
\begin{equation}
    \hat{\textbf{x}}=M \odot \textbf{x},
\end{equation}
where $\odot$ is the element-wise product operation.  
As in \citep{pathak2016context_encoders_inpainting}, we use a two-terms loss function. Following the same mathematical notation, the first term is called the \textit{reconstruction loss} $\mathcal{L}_{rec}$ and considers the masked part of the patch. The model $F$ learns by minimizing the L$2$-norm between the prediction of the masked area and the original masked part. It is expressed as:
\begin{equation}
    \mathcal{L}_{rec}(\hat{x})=\frac{1}{\sum (1-M)} \lVert(1-M)\odot (x-F(\hat x))\rVert_2^2.
\end{equation}
Differently, the second term $\mathcal{L}_{con}$ concentrates on the reconstruction of the unmasked part present in the input, which is responsible for the spatial context. It can be expressed as:
\begin{equation}
    \mathcal{L}_{con}(\hat{x})=\frac{1}{\sum (M)} \lVert M\odot (x-F((1-M)\odot x))\rVert_2^2.
\end{equation}
Finally, we build the inpainting loss function  $\mathcal{L}_{inpainting}$ as: 
\begin{equation}
    \mathcal{L}_{inpainting} = w_{rec}\mathcal{L}_{rec} + (1-w_{rec})\mathcal{L}_{con},
\end{equation}
with the weighting coefficient $w_{rec}=0.99$ as proposed in \citep{singh2018ssl_semantic_segmentation_overhead_imagery}, where 
it was shown to be a good balance between inpainting and learned feature quality. This emphasis on learning from the masked area is crucial because this area represents missing or corrupted information in the input data. By prioritizing the learning from these regions, the model can effectively understand the context, structure and relationships of the surrounding data points.

\subsection{Forest mapping downstream task} 
\label{sssec:sl_model}

We rely on the U-Net model introduced in Section~\ref{ssec:cnn_architectures} for the downstream task of forest mapping (\mbox{Figure~\ref{fig:dl_architectures}(b)}).
Note that we distinguish two different uses of the U-Net, depending on the initialization of the encoder weights:
when a random initialization is considered, the U-Net is trained in a classic FSL manner. 
Differently, when we transfer knowledge from the CAE trained using an SSL task, we refer to the use of the U-Net as DST. In the following, we identify these two cases as FSL and DST, respectively.

In our study, the definition of the loss function for training the U-Net is conditioned by the challenge imposed by the many \mbox{TanDEM-X} acquisition geometries and the class-imbalance between 
forest and non-forest samples among the input patches.
Following the findings in \citep{Jadon2020loss_functions_semantic_segmentation} we select a combination of the Binary Cross Entropy (BCE) loss and the dice loss. 
The $\mathcal{L}_{BCE}$ is defined as a measure of the difference between two probability distributions for a given random variable or set of events. It is widely used for classification purposes at pixel level and can be expressed as:
\begin{equation}
\label{eq:l_bce}
    \mathcal{L}_{BCE}(y,\hat{y}) = -(y \cdot \mathrm{log}(\hat{y})+(1-y) \cdot \mathrm{log}(1-\hat{y})),
\end{equation}
where $y$ represents here the binary reference label, typically labeled as 0 (negative class) and 1 (positive class), and $\hat{y}$ denotes the predicted probability to belong to the positive class.

On the other hand, the dice loss $\mathcal{L}_{DICE}$ is derived from the dice coefficient \citep{dice}, which is widely used in computer vision to compute the similarity between two images
and can be computed as:
\begin{equation}
\label{eq:l_dice}
    \mathcal{L}_{DICE}(y,\hat{y}) = 1- \frac{y \cdot \hat{y}+1}{y+\hat{y}+1}.
\end{equation}

\noindent 
where 1 is added in numerator and denominator as regularization terms to 
avoid divisions \mbox{by 0} when $y=\hat{y}=0$. 
The resulting loss function for the FSL and DST cases is finally given by:

\begin{equation}
    \mathcal{L}_{\mathrm{U\textrm{-}Net}} = \mathcal{L}_{BCE} + \mathcal{L}_{DICE}.
\end{equation}

\subsection{Training strategies}
\label{sseq:training_strategy}

In this section, we summarize the details of the different training phases of all the investigated configurations. 

\subsubsection{Fully-supervised learning}
\label{sssec:fsl_training}

\begin{figure}
    \centering
    \includegraphics[width=0.95\textwidth]{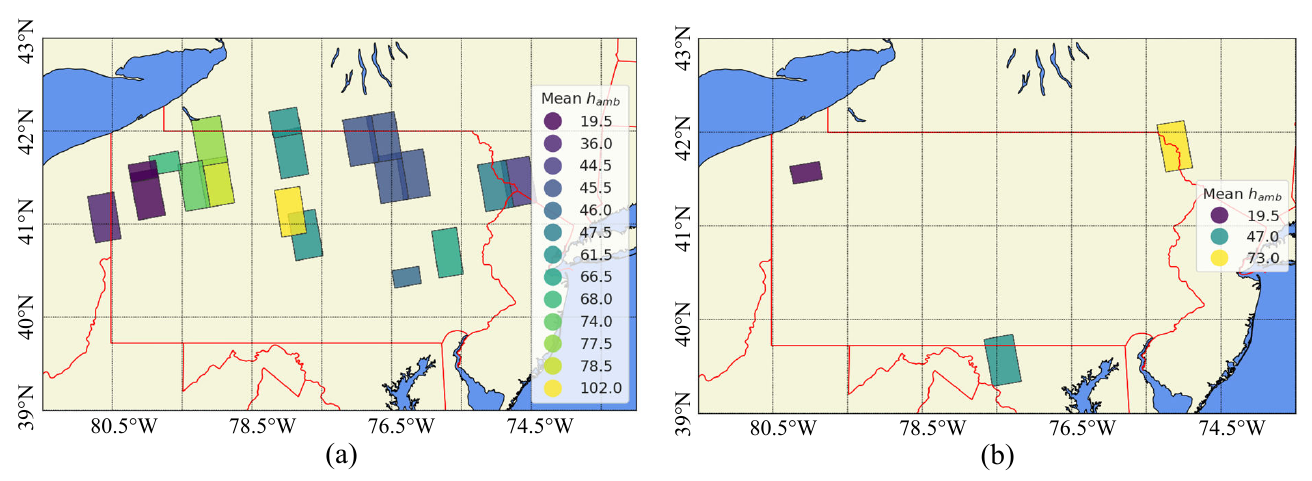}
    \caption{Considered \mbox{TanDEM-X} acquisitions when using (a) 22\% and (b) 1.5\% of the available labeled data over Pennsylvania. They are used for training the downstream task of forest mapping.}
    \label{fig:competitive_scenarios}
\end{figure}

As baseline scenario, the U-Net presented in this study can be trained in a fully-supervised manner to directly perform the forest mapping task.
We consider 450 \mbox{TanDEM-X} images acquired between 2011 and 2012 over the Pennsylvania state, USA (\mbox{Figure~\ref{fig:data}}(a)).
This represents an ideal scenario of using \textbf{100\%} (103,500km$^2$) of reference labeled data.
To simulate a real-case scenario, where not so many reference data are available, we also train the U-Net in a fully-supervised way simply reducing the amount of used labeled data.
We consider three cases using 
\textbf{1.5\%} (1,450km$^2$),
\textbf{8\%} (8,100km$^2$)
and \textbf{22\%} (22,800km$^2$)
of the available input data, respectively.
In all cases we select \mbox{TanDEM-X} images representative of the global variability of $h_{\mathrm{amb}}$, as can be observed in \mbox{Figure~\ref{fig:competitive_scenarios}}. This figure represents the considered \mbox{TanDEM-X} acquisitions when using 1.5\% and 22\% of the available labeled data.

For the validation of the different training scenarios, 
we use the same \mbox{TanDEM-X} images in all cases.

\subsubsection{Self-supervised learning for the pretext task}
\label{sssec:ssl_training}

To train our CAE model in SSL manner we use 1500 \mbox{TanDEM-X} images acquired from 2010 up to 2022 over the Pennsylvania state and its neighborhood area, green and brown acquisitions depicted in \mbox{Figure~\ref{fig:data}(a)} and described in \mbox{Section~\ref{sec:materials}}. 
We pay special attention to have a balanced number of forest and non-forest samples.
To account for the different acquisition
geometries in the \mbox{TanDEM-X} mission, the considered $h_{\mathrm{amb}}$ values range are divided in intervals
of 2~m and for each interval, up to 20 \mbox{TanDEM-X} images are used for training and 10 for validation.
Finally, we consider \mbox{TanDEM-X} acquisitions in ascending and descending orbit directions as well.

\subsubsection{Forest mapping downstream task}
\label{sssec:sl_training}

To assess the influence of the SSL pre-training on forest mapping with \mbox{TanDEM-X} data at 6~m resolution, we carry out different experiments over the Pennsylvania test region.
The settings of the experiments vary depending on different training approaches:

\begin{itemize}
\item \textit{U-Net weights initialization:} 
The U-Net encoder weights can be initialized with the weights obtained after SSL pre-training.
We distinguish between the encoder weights obtained after the two investigated SSL pretext tasks as 
\textbf{SSL-Id} and \textbf{SSL-In}.

\item \textit{U-Net trainability:} After encoder weights initialization, 
we have two possibilities to train the U-Net: to train both encoder and decoder 
or to freeze the weights coming from the SSL model and to train only the decoder part of the U-Net, which in both cases is randomly initialized.
In our study we identify these cases as (\textbf{E+D}) when training both encoder and decoder 
and (\textbf{D}) when training only the decoder. 
\item \textit{Amount of reference labeled data used:} The number of input patches can be varied to assess 
the influence of the SSL pre-training on the final forest mapping performance. 
In this study (as for the case of fully-supervised training cases, \mbox{Section~\ref{sssec:fsl_training}}), 
we selected three cases using 
\textbf{1.5\%}, 
\textbf{8\%}, 
and \textbf{22\%} 
of the input available data
accounting for a proper representation of the $h_{\mathrm{amb}}$ and the land cover classes.
The same validation strategy as for FSL case is followed.
\end{itemize}

\subsection{Testing strategy}
\label{sseq:val_test_strategy}

For testing purposes of all training scenarios, 
we consider a testing region characterized by the presence of different land cover classes.
It confines with the Pennsylvania state borders on the South-West corner and ranges in latitude from 
39.72$^{\circ}$N to 40.75$^{\circ}$N and in longitude from 77.98$^{\circ}$W to 80.52$^{\circ}$W.

For the purpose of testing across the whole 
$h_{\mathrm{amb}}$ range, we build different sets of test acquisitions of similar sizes in which we distinguish the following ranges 
of $h_{\mathrm{amb}}$ values: short ($h_{\mathrm{amb}} < 40~\mathrm{m}$), mid ($h_{\mathrm{amb}} \in [40~\mathrm{m} - 60~\mathrm{m}]$) 
and large ($h_{\mathrm{amb}} > 60~\mathrm{m}$). 
We use only \mbox{TanDEM-X} images acquired during 2011 and 2012, in order to reduce the time lag with respect to the reference map and to increase the reliability of the evaluation. 

\begin{table}[]
\centering
    \caption{Details of defined testing subsets over Pennsylvania.}
    \label{tab:test_subsets}
\resizebox{0.75\columnwidth}{!}{
\begin{tabular}{cccccc} \hline
$h_\mathrm{amb}$ range & Orbit dir. & Nr. patches & Nr. pixels & Px. forest & Px. non-forest \\
 (m) &  & & &   (\%) &  (\%) \\ \hline
20 - 40 & Ascending & 59,991 & 982,892,544 & 62 & 38 \\
40 - 60 & Ascending & 46,098 & 755,269,632 & 61 & 39 \\
60 - 120 & Ascending & 38,728 & 634,519,552 & 61 & 39 \\
40 - 90 & Descending & 51,084 & 836,960,256 & 60 & 40 \\ \hline
\end{tabular}
}
\end{table}

To evaluate the impact of the SSL learning on the final downstream task 
and on the generalization capabilities of the different DL approaches, 
we include a fourth subset with \mbox{TanDEM-X} images acquired in 2013 in descending orbit over the test region. 
\mbox{Table~\ref{tab:test_subsets}} shows the details for each test subset.

We train and test three times each combination of DL approaches and test subsets to mitigate how randomness affects the final results. 
The presented results correspond to the average obtained after the corresponding runs of each combination of the F$^w_1$-score values.

Clearly, in order 
to avoid information leakage,
the test images are never used for any learning task.

\subsubsection{Performance metrics}
\label{ssec:metrics}
We assess the quality our results using the following standard performance metrics: 
overall accuracy, precision, recall and $F_1$-score.
Starting from the definition of confusion matrix \citep{townsend1971theoretical}, we can assess the result of the classification for each pixel in terms of: true positives (TP) - pixels
correctly classified as forest, false positives (FP) -  pixels predicted as forest but labeled as non-forest in the reference map, false negatives (FN) - pixels predicted as non-forest and labeled as forest in the reference map, and true negatives
(TN) - pixels correctly classified as non-forest in both maps. 

The Overall Accuracy (OA) represents 
the ratio between true predictions (positive and negative) and the total number of observations. It is computed as:
\begin{equation} 
\label{eq:accuracy}
\textrm{OA} = \frac{TP + TN}{TP + TN + FP + FN}.
\end{equation}

\noindent However, the overall accuracy is prone to biases introduced by class imbalance, in which case an accuracy assessment based on precision, recall, and $F_1$-score should be preferred. 
Indeed, precision (also known as user’s
accuracy) is the ratio between true positives and the number of all positive predictions:
\begin{equation}
\textrm{Precision} = \frac{\textrm{TP}}{\textrm{TP} + \textrm{FP}}.
\label{eq:precision}
\end{equation}
On the other hand, recall (also known as producer's accuracy or sensitivity) represents the probability of detection, i.e., the proportion of positives that are correctly identified:
\begin{equation}
\textrm{Recall} = \frac{\textrm{TP}}{\textrm{TP} + \textrm{FN}}.
\label{eq:recall}
\end{equation}
Finally, precision and recall are often combined together as they represent the balance between capturing the target class whenever it appears (FN = 0, Recall = 1) and minimizing false alarms (FP = 0, Precision = 1). A concise representation of the two measures is their harmonic average, known as the F$_1-$score, which is more robust in presence unbalanced datasets and is defined as:
\begin{equation}
\textrm{F}_1\mathrm{-score} = 2 \frac{\textrm{Precision} \cdot  \textrm{Recall}}{\textrm{Precision} + \text{Recall}}. 
\label{eq:f1}
\end{equation}

For the classification of class-imbalanced data, we rely on the computation of the weighted F$_1-$score (F$_1^w-$score), which represents an averaged unique value of the i class-wise F$_1-$scores using as weights the number of samples from each class. It can be expressed as:
\begin{equation}
\textrm{F}_1^w\mathrm{-score} = \sum_{i=1}^{N} w_i \textrm{F}_1\mathrm{-score}_i,
\label{eq:weighted_f1}
\end{equation}

\noindent where $N$ represents the number of considered land cover classes and the weight $w$ for each class $i$ is calculated as $w_i = N_i/N$, with $N_i$ identifying the number of samples of class $i$. 

\section{Experiments and results}
\label{sec:results}

\subsection{Experiments over the Pennsylvania temperate forest}
\label{ssec:results_penn}

\begin{figure}
    \centering
    \includegraphics[width=0.82\textwidth]{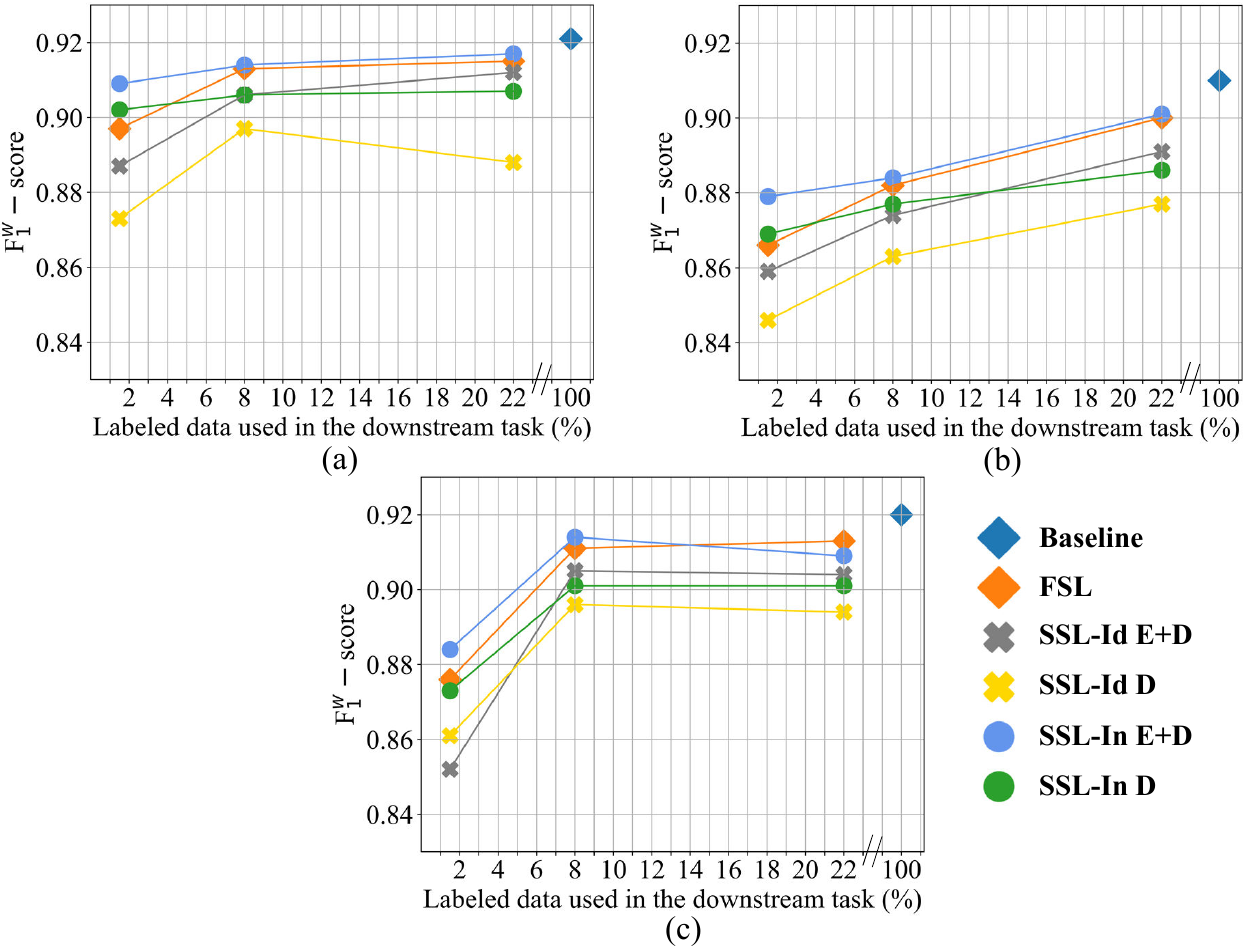}
    \caption{$F^w_1$-score for the investigated approaches over Pennsylvania testing area for the test subsets:
    (a) Short $h_\mathrm{amb}$;
    (b) Mid $h_\mathrm{amb}$;
    (c) Large $h_\mathrm{amb}$;    
     The results are presented for the different amount of considered labeled data in the supervised learning part: 1.5\%, 8\% and 22\% (horizontal axis).
     The \textbf{baseline} approach consists of a fully-supervised training using 100\% of the available \mbox{TanDEM-X} images with labeled data. \textbf{FSL} corresponds to the same architecture of the baseline (U-Net) trained in a fully-supervised manner with less labeled data. The other cases correspond to SSL pretext task (\textbf{SSL-Id}: identity reconstruction with CAE, \textbf{SSL-In}: inpainting with masked CAE), followed by a downstream task of forest mapping (\textbf{E+D}: encoder and decoder initialized from the SSL pretext task and then trained, \textbf{D}: encoder weights frozen from SSL pretext task and decoder trained). 
     }
    \label{fig:f1_vs_perc_input_samples}
\end{figure}

The results obtained for all the experiments with the dataset reserved for testing purposes over the Pennsylvania temperate forest are presented in \mbox{Figure~\ref{fig:f1_vs_perc_input_samples}}. They are separately shown for each defined test $h_{\mathrm{amb}}$ interval (as discussed in \mbox{Section~\ref{sseq:val_test_strategy}}).
The results represent the F$_1^w$-score. 
We present the complete results in \ref{appendix}.
We select the best performing model in each case for the further classification of \mbox{TanDEM-X} images into forest/non-forest maps.

As expected, the \textbf{baseline} scenario using 100\% of the available \mbox{TanDEM-X} patches and training the U-Net in a fully-supervised manner achieves the best results independently of the $h_{\mathrm{amb}}$ range.
The performance then decreases when considering a smaller dataset for fully-supervised training (\textbf{FSL} case).

Regarding the SSL pretext tasks + downstream forest mapping task, we can see how the identitiy reconstruction (\textbf{SSL-Id}) always presents a worse performance than inpainting (\textbf{SSL-In}). Regarding this last SSL task, the subsequent knowledge transfer to the U-Net for the downstream task achieves a better performance when initializing the U-Net weights with the ones from the CAE and then training both encoder and decoder (\textbf{SSL-In E+D}) rather than when fixing the encoder and training the decoder only (\textbf{SSL-In D}). Specifically, the  \textbf{SSL-In E+D} case shows a very competitive performance with respect to the fully-supervised case, particularly when significantly reducing the amount of training patches, which confirms the 
 meaningfulness of the features learned by the pretext task.

\begin{figure}
    \centering
    \includegraphics[width=0.8\textwidth]{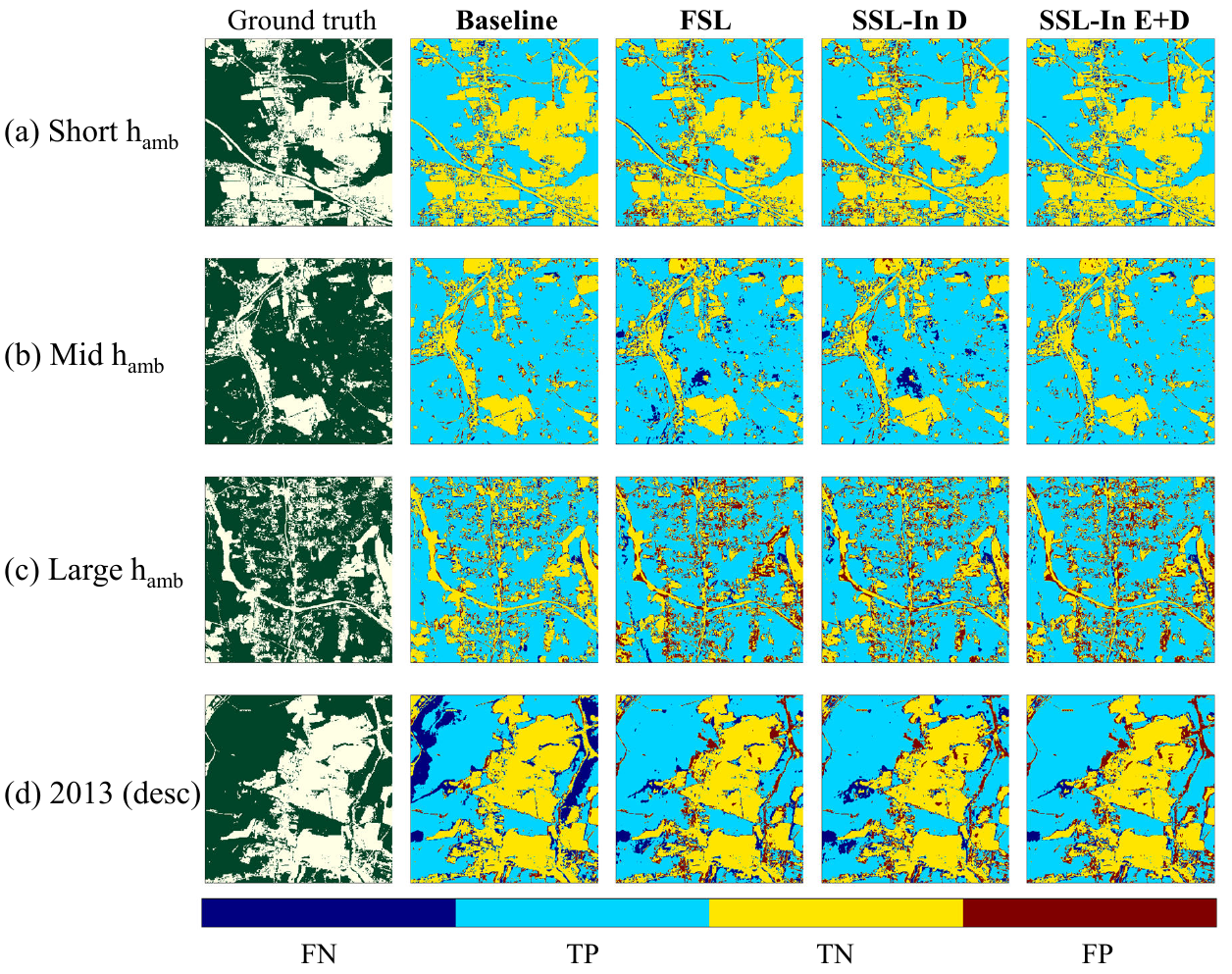}
    \caption{Map view of the confusion matrices values for 4 different areas on patches of 1024 $\times$ 1024 pixels. On the ground truth plots on the left-hand side, green areas correspond to forests and white areas to non-forested zones. Beside the \textbf{baseline} case, the models are trained with 1.5\% of the labeled data. The rows correspond to areas inside images acquired in ascending orbit direction in 2011 and 2012 for the different test subsets: (a) Short $h_{\textrm{amb}}$, (b) Mid $h_{\textrm{amb}}$, and (c) Large $h_{\textrm{amb}}$. The fourth row (d) is part of a \mbox{TanDEM-X} image acquired in 2013 in descending orbit direction and with a $h_{\textrm{amb}} = 85~m$.}
    \label{fig:class_visual}
\end{figure}

Therefore, in light of these results, the case \textbf{SSL-In E+D} represents the best DL model over the Pennsylvania test region in a real-case scenario with just 1.5\% of available labeled training data.

A visual comparison of the different DL model performance using 1.5\% of the labeled data for training is presented in \mbox{Figure~\ref{fig:class_visual}}, which 
represents the components of the confusion matrix (TP, FP, TN, and FN) for different patches of 1024 $\times$ 1024 pixels. The corresponding quantitative results in terms of F$^w_1-$score are presented in Table~\ref{tab:visual}.
Each row corresponds to a different \mbox{TanDEM-X} testing image (different $h_{\mathrm{amb}}$ ranges), while each  column from left to right depicts the reference ground truth, the \textbf{baseline} case, the \textbf{FSL} case using only 1.5\% of the labeled data, the \textbf{SSL-In D} case and, finally, the best performing \textbf{SSL-In E+D} case.
We can observe that all the networks correctly classify large forested and non-forest areas. In general, the small differences in performance between the networks mainly lie in their ability to correctly classify the delimitation zones between forest and non-forest. 
Due to the side looking geometry of \mbox{TanDEM-X} InSAR acquisitions and the very high resolution of TanDEM-X, forest borders often appear as shadowed areas (characterized by both low backscatter and coherence), leading to an overestimation of forests. This error is particularly notable around roads passing through a forest.
When considering the \textbf{SSL-In E+D} case, comparable results to the \textbf{baseline} are obtained.
The improvement in the classification is especially observable in the \mbox{TanDEM-X} images acquired with a different orbit direction, only seen in the training of the SSL model. This is the case of the \mbox{TanDEM-X} images acquired in 2013 in descending orbit direction. 
Indeed, we observe some misclassification in the \textbf{baseline} case trained with all \mbox{TanDEM-X} images acquired in 2011 and 2012, which were acquired in ascending orbit direction only. This fact emphasizes the importance of training any DL approach applied to SAR images with all possible acquisition geometries and orbit directions.

\begin{table}[t]
\centering
\resizebox{0.6\columnwidth}{!}{
\begin{tabular}{lcccc}\hline
\multicolumn{1}{c}{Row} &  \textbf{Baseline} & \textbf{FSL} & \textbf{SSL-In D} & \textbf{SSL-In E+D} \\[10pt]\hline
(a) Short $h_{\textrm{amb}}$& {0.916} & 0.901 &0.89& {0.906} \\[10pt] \hline
(b) Mid $h_{\textrm{amb}}$& {0.952} & 0.928 &0.92 & {0.948} \\[10pt] \hline
(c) Large $h_{\textrm{amb}}$& {0.888} & 0.851 &0.85 & 0.858 \\[10pt] \hline
(d) 2013 (desc) & 0.799 & 0.852 & 0.846 & {0.864} \\[12pt] \hline
\end{tabular}
}
\caption{F$^w_1$-score associated to the results in Figure \ref{fig:class_visual}. The rows correspond to the different test subsets \mbox{({Section~\ref{sseq:val_test_strategy}})} and the columns to different investigated DL approaches.}
\label{tab:visual}
\end{table}

\begin{figure}
    \centering
    \includegraphics[width=1\textwidth]{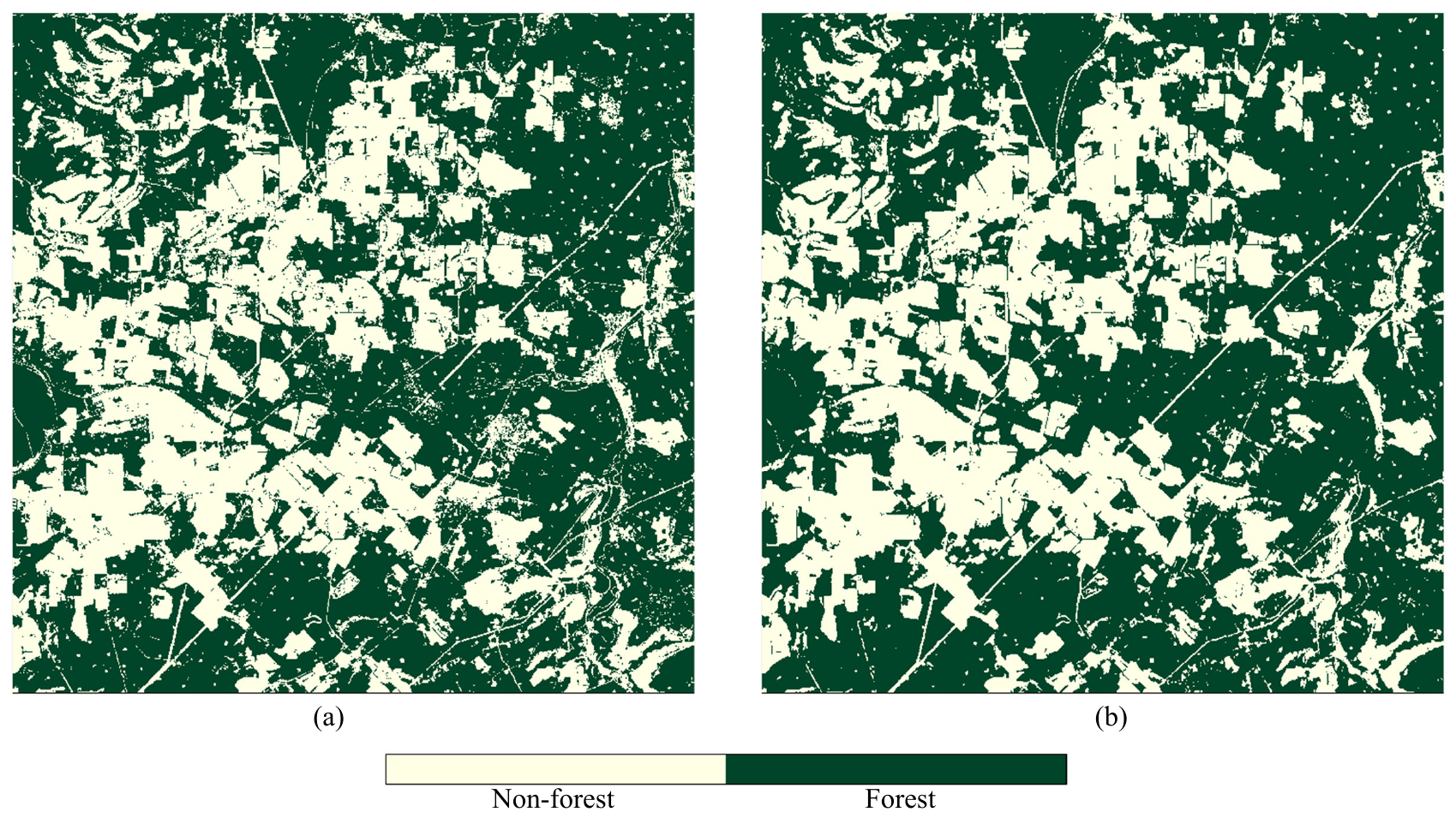}
    \caption{Comparison between the ground truth (a) and the predicted segmentation (b) of a 2048 pixels $\times$ 2048 pixels patch. \mbox{TanDEM-X} acquisition taken with a $h_{\textrm{amb}}=44~\mathrm{m}$ in 2011 over the DL testing region.
    Small clear-cut areas ("point-like" with a diameter of approx. 30~m) can be noticed especially in the lower left corner and the top-right corner of the images.}
    \label{fig:patch}
\end{figure}

As a final example of the best performing model when using only 1.5\% of the available \mbox{TanDEM-X} images over Pennsylvania temperate forest, in \mbox{Figure~\ref{fig:patch}} we compare the ground truth and the \textbf{SSL-In E+D} classification for an image crop of 2048 $\times$ 2048 pixels. The \mbox{TanDEM-X} image was acquired in 2011 with a $h_{\mathrm{amb}} = 44~\mathrm{m}$ over the testing region.
Although some details are lost, most of the roads and paths, even through dense forest, are well detected. 
Especially noticeable are the correct detection and delimitation of small clear-cut areas with a definite geometric closed shape. They show circular patterns, with an approximate diameter of 30~m and are mostly located in the lower left as well as upper right corner of the figures.

\subsection{Application scenario over the Amazon rainforest}
\label{ssec:app_scenario}

In light of our findings over temperate forests,
we now apply the best performing DL approach (i.e., \textbf{SSL-In E+D}) to classify \mbox{TanDEM-X} data takes acquired over the Amazon rainforest, where few input samples are available as reference. 
The re-training of the DL model is necessary due to different type of forest, as done by \citep{martone2018FNF}.
Moreover, for performance comparison purposes, we also train the U-Net in a fully-supervised manner with the available reference data (\textbf{FSL} case).
\mbox{Table~\ref{tab:patches_ama}} shows the details of the used LiDAR patches for training and validation. We select them to assure a certain balance between forest and non-forest samples, with mainly all the patches having a forest coverage between 40\% and 80\%. The imaged area of the patches ranges from 1 to 10 $\mathrm{km}^2$. 

For the SSL part, we train the CAE using a selected set of \mbox{TanDEM-X} InSAR acquisitions (\mbox{Section~\ref{ssec:tdx_dataset}}).
We use 480 \mbox{TanDEM-X} images for training and 190 for validation.  
To account for the different acquisition geometries in the \mbox{TanDEM-X} mission, we divide the considered $h_{\mathrm{amb}}$ values range 
in intervals of 2~m and for each interval, we use up to 10 \mbox{TanDEM-X} images for training and 5 for validation.

After proper transfer of the encoder weights from the CAE to the U-Net, we train
both encoder and decoder using the few available labeled patches (\mbox{Section~\ref{ssec:ref_data_ama}}). 
As input dataset, we consider \mbox{TanDEM-X} images acquired within a time span of $\pm 1$ years with respect to the reference patches. 
These \mbox{TanDEM-X} images were acquired with different acquisition geometries over the Pará state and overlap the available LiDAR reference patches presented in \mbox{Table~\ref{tab:patches_ama}}.

\begin{table}[t]
\centering
    \caption{Acquisition details of the Forest/Non-forest patches used as reference data to test the proposed SSL approach over the Amazon rainforest. }
    \label{tab:patches_ama}
\resizebox{0.8\columnwidth}{!}{
\begin{tabular}{ccccccc}
\hline
Patch Nr. & Acq. Year & Center coord. & Extension & Forest & Non-forest & Original point\\
 & & (lat, lon) & (pixels, lat $\times$ lon) & (pixels)  & (pixels) & cloud dataset\\
 \hline
1 & 2012      & (-3.13, -54.95)           & 922 $\times$ 329                     & 146,640 & 40,703 & TAP\_A03\_2012      \\ 
2 & 2012      & (-6.41, -52.90)           & 863 $\times$ 439                     & 193,207 & 84,517 & SFX\_A01\_2012     \\ 
3 & 2013      & (-2.46, -48.31)           & 834 $\times$ 807                     & 181,506 & 93,751 & TAC\_A01\_2013     \\ 
4 & 2013      & (-2.98, -46.90)           & 373 $\times$ 776                     & 18,836  & 10,553  & PRG\_A01\_2013\_P02a    \\ 
5 & 2013      & (-3.05, -47.06)           & 373 $\times$ 776                     & 11,413  & 18,002 & PRG\_A01\_2013\_P05b     \\ 
6 & 2013      & (-3.03, -47.03)           & 373 $\times$ 776                     & 12,512  & 16,899 & PRG\_A01\_2013\_P05a     \\ 
7 & 2013      & (-3.06, -47.46)           & 374 $\times$ 776                     & 1,776   & 27,641 & PRG\_A01\_2013\_P10a     \\ 
8 & 2013      & (-3.11, -46.81)           & 372 $\times$ 775                     & 19,706  & 9,613 & PRG\_A01\_2013\_P03a      \\ 
9 & 2013      & (-3.12, -54.98)           & 489 $\times$ 271                     & 71,540  & 23,808 & TAP\_A02\_2013     \\ 
10 & 2013      & (-3.13, -54.95)           & 921 $\times$ 329                     & 138,201 & 48,967 & TAP\_A03\_2013     \\ 
11 & 2013      & (-3.32, -47.30)           & 374 $\times$ 776                     & 23,007  & 6,355 & PRG\_A01\_2013\_P12b      \\ 
12 & 2016      & (-3.13, -54.95)           & 921 $\times$ 330                     & 122,233 & 65,307 & TAP\_A03\_2016      \\ 
13 & 2017      & (-12.25, -55.10)          & 289 $\times$ 351                     & 40,633  & 18,474 & FND\_A01\_2017     \\ 
14 & 2018      & (-2.50, -54.66)           & 605 $\times$ 622                     & 23,286  & 6,074 & ST3\_A01\_2018\_P03      \\ 
15 & 2018      & (-2.60, -54.54)           & 784 $\times$ 348                     & 22,097  & 7,141  & ST3\_A01\_2018\_P05   \\   \hline
\end{tabular}
}
\end{table}

\begin{figure}
    \centering
    \includegraphics[width=0.7\textwidth]{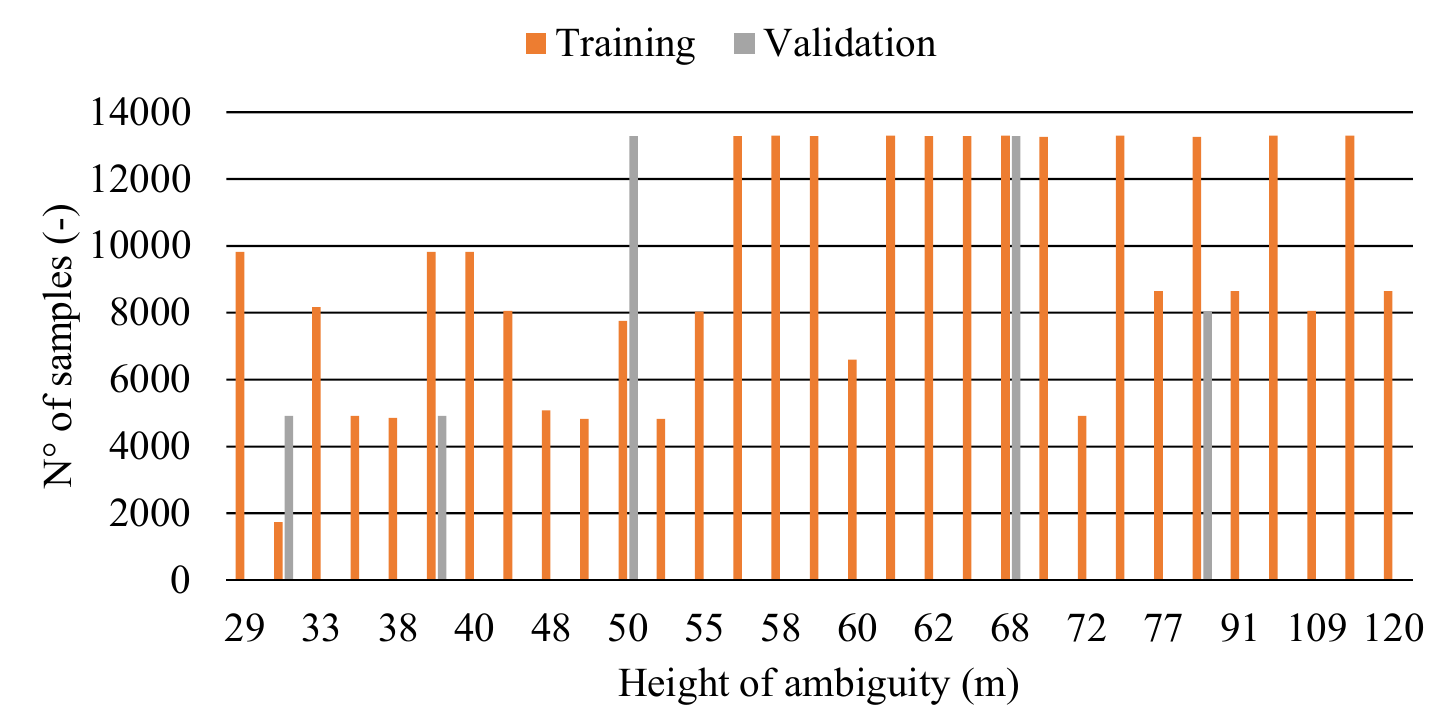}
    \caption{Number of pixels used for training and validation of the supervised downstream forest mapping task over the Amazon rainforest.}
    \label{fig:train_val_amazonas}
\end{figure}

\begin{figure}
    \centering
    \includegraphics[width=0.9\textwidth]{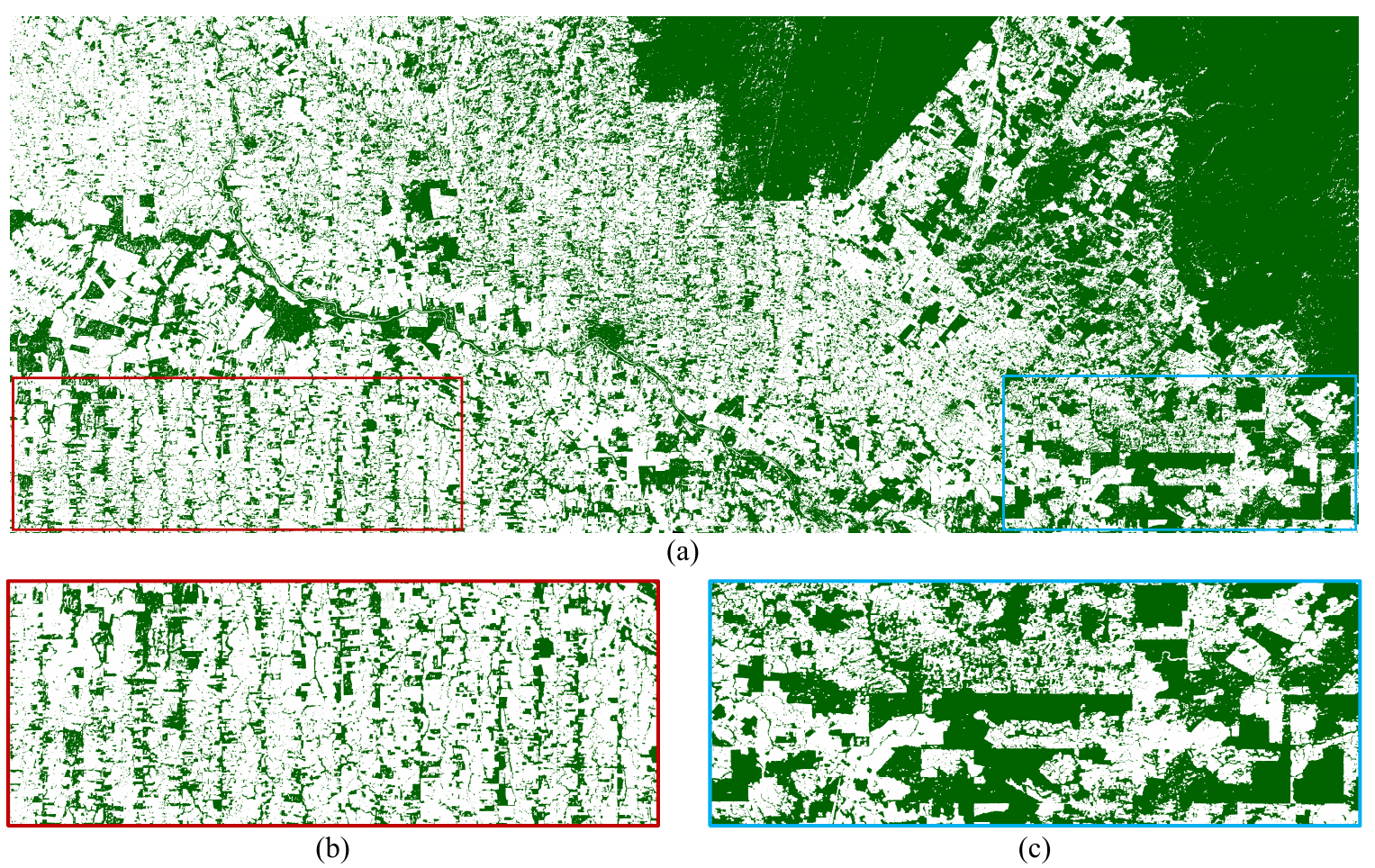}
    \caption{(a) Mosaic over Rondônia state using \mbox{TanDEM-X} InSAR data at 6~m resolution. (b) and (c) highlight two zoomed-in areas of (a). Forested and non-forested areas are indicated in green and white, respectively.}
    \label{fig:mosaic_rondonia}
\end{figure}

\mbox{Figure~\ref{fig:train_val_amazonas}} shows the distribution of the training and validations subsets
for the supervised downstream task over the Amazon rainforest. We look for an homogeneous distribution of the
used samples over the whole $h_{\mathrm{amb}}$ range.

To evaluate the performance of the proposed SSL DL approach over this challenging area, 
we classify more than 500 \mbox{TanDEM-X} images acquired in 2019 and 2020 over the South-East region of the Amazon rainforest using both the fully-supervised \textbf{FSL} model and the proposed \textbf{SSL-In E+D} approach. Finally, we downsample the generated forest/non-forest maps to 10~m, in order to be
intercompared with the forest map derived from the ESA CCI HRLC map (Section~\ref{ssec:esa_hrlc_cci}).

\mbox{Figure~\ref{fig:mosaic_rondonia}} shows a mosaic of forested areas over the Rondônia state, Brazil (with longitude range 60.5$^\circ$W - 62.25$^\circ$W, and latitude range 11.0$^\circ$S - 11.7$^\circ$S) obtained with \mbox{TanDEM-X} data at 6~m resolution. We show the details of two crop areas.
We observe, that typical clear-cuts patterns over the Amazon rainforest are well detected (\mbox{Figure~\ref{fig:mosaic_rondonia}(b)}) and larger forested areas are well delimited \mbox{Figure~\ref{fig:mosaic_rondonia}(c)}.
This mosaic is a crop of the generated map used for intercomparison of our results with the ESA CCI HRLC map. \mbox{Table~\ref{tab:amazonas_performance}} depicts the performance obtained after maps intercomparison for the forest and non-forest classes. 

The overall accuracy improves by 14\% when using the \textbf{SSL-In E+D} approach with respect to the fully-supervised \textbf{FSL}, which suffers from the lack of extended reference data sets.
For the analysis of the non-forest class and in order to avoid evaluating the trivial cases of predominantly vegetated scenes, we present two $F_1$-score values for images with a proportion of non-forest pixels larger than 5\% and 10\%, respectively.
Regarding the \textbf{FSL} case, we observe a tendency to overestimate forested areas, while a significant improvement in performance is confirmed when considering the \textbf{SSL-In E+D} case, reaching values higher than 0.7. 
Taking a closer look to the precision and recall of the forest class, we observe a very high precision with both DL approaches, while the \textbf{SSL-In E+D} case achieves a higher recall. 

\begin{table}[t]
\centering
    \caption{Classification performance obtained over the Amazon rainforest after comparing the predicted \mbox{TanDEM-X} forest/non-forest maps with the ESA CCI HRLC map at 10 m resolution. LC class refers to the predicted class (either forest or non-forest) and DTs is the number of considered \mbox{TanDEM-X} images. For the non-forest class, we consider only images with an amount of non-forest pixels higher than 5\% and 10\%, respectively.}
    \label{tab:amazonas_performance}
\resizebox{0.55\columnwidth}{!}{
\begin{tabular}{ccccc}
\hline
Parameter  & LC class &  DTs & \textbf{FSL}   & \textbf{SSL-In E+D}  \\ \hline
Accuracy         & All & 505     & 0.65 & 0.74 \\ 
$\textrm{F}_1$-score & Non-Forest (5\%) &  346     & 0.48 & 0.66 \\ 
$\textrm{F}_1$-score & Non-Forest (10\%) & 295     & 0.47 & 0.72 \\ 
$\textrm{F}_1$-score & Forest &  505     & 0.62 & 0.77 \\ 
Precision     & Forest &  505     & 0.94 & 0.94 \\ 
Recall        & Forest & 505     & 0.54 & 0.68 \\ \hline
\end{tabular}
}
\end{table}

\begin{figure}
    \centering
    \includegraphics[width=0.95\textwidth]{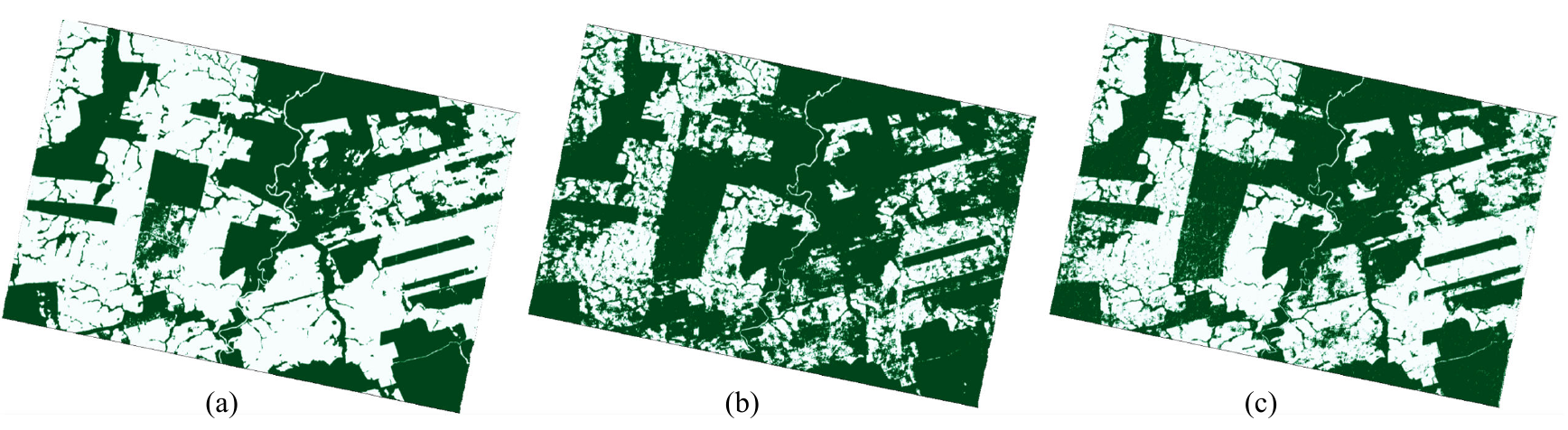}
    \caption{Maps intercomparison over the Amazon rainforest at 10 m: (a) Forest layer derived from the ESA CCI HRLC map; (b) Forest classification using a \mbox{TanDEM-X} acquisition and the \textbf{FSL} approach; (c) Forest classification of a \mbox{TanDEM-X} image and \textbf{SSL-In E+D}.}
    \label{fig:intercomparison_amazonas}
\end{figure}

As an example, \mbox{Figure~\ref{fig:intercomparison_amazonas}} shows the classification results of a \mbox{TanDEM-X} image zoom-in acquired over the Amazon rainforest. This image has been acquired with a $h_{\mathrm{amb}} = 69~\mathrm{m}$ and has as center coordinates 7.94$^\circ$S and 55.33$^\circ$W. Two classifications are presented and compared with the forest detected by the ESA CCI HRLC map, shown in \mbox{Figure~\ref{fig:intercomparison_amazonas}(a)}.
\mbox{Figure~\ref{fig:intercomparison_amazonas}(b)} and \mbox{Figure~\ref{fig:intercomparison_amazonas}(c)} depict the classification obtained with the \textbf{FSL} and the \textbf{SSL-In E+D}, respectively.
With the \textbf{FSL} approach we obtain a $F_1$-score
of 0.79 for the forest class and 0.73 for the non-forest areas, while with the \textbf{SSL-In E+D} classification we achieve similar $F_1$-score of 0.90 for both land cover classes.

\section{Discussion}
\label{sec:discussion}

The results presented in this analysis demonstrate the added value of the \mbox{TanDEM-X} InSAR dataset for the detection of forested areas at very high resolution. The volume correlation factor, directly derived from the bistatic coherence, adds valuable information to the backscatter signal typically recorded by monostatic SAR systems.
The possibility to interferometrically process the \mbox{TanDEM-X} dataset at 6~m independent pixel spacing allows for a precise delineation of forested areas, which is crucial to detect narrow paths and rivers in between dense forested areas, as well as to estimate clear-cuts in the order of a few tens of square meters.
As it is shown in \mbox{Figure~\ref{fig:class_visual}}, most of the classification errors (FN and FP) mainly occur in correspondence of the borders of the forested areas. This is due to the side-looking geometry of SAR, which leads to the presence of shadowed regions, clearly visible at high resolution.
Our analysis indicates that the distinguishing factor in terms of performance among the investigated DL methods is their ability to accurately delineate the border between forested and non-forested areas, rather than the general detection of forested areas, which is accurately performed in all cases.
In this regard, the acquisition of 
such areas with different orbit direction, as done for the generation of the \mbox{TanDEM-X} global DEM over mountainous terrain \citep{rizzoli2016DEMperformance}, would help improving the classification.

The application of DL methods for forest mapping at such a fine resolution suffers from the lack of reference data, and
SSL represents an alternative to overcome this limitation.
The capability of autoencoders to extract and learn the most important features contained in the input dataset, make them very suitable to be applied with InSAR data.
In our study, already during the SSL training 
using both pretext tasks, we observe some patterns related to forest. Non-forested areas are quite homogeneous while forested areas present some roughness.
This behavior helps the posterior downstream task for forest mapping by properly pointing the U-Net towards a correct final solution,
highlighting the impact of SSL on the starting point for the supervised DST.

One key finding from our SSL experiments is the demonstration of the effectiveness of the pretext task for dealing with the lack of extended reference data sets. 
In particular, the identity pretext tasks does not provide any noticeable benefit in the context of our forested area detection task. As depicted in \mbox{Figure~\ref{fig:f1_vs_perc_input_samples}}, we see that even with very few labeled data, a random initialization of the U-Net (FSL) shows better results.
On the other hand, the inpainting pretext task helps the autoencoder to properly learn suitable representations from the input InSAR dataset.
Indeed, after transferring the weights learned by the autoencoder with the inpainting pretext task to the U-Net and training both, encoder and decoder, this DL approach
improves the performance and convergences of the downstream forest mapping task with respect to the other analyzed learning methods.
Furthermore, we observe that pre-training using an inpainting task improves the final performance on \mbox{TanDEM-X} data used only for SSL training.
As it is shown in \mbox{Figure~\ref{fig:class_visual}} and the patch acquired in 2013 in descending orbit, the SSL-In approach outperforms the DL approach using all 2011 and 2012 \mbox{TanDEM-X} images with labeled data and acquired in ascending orbit.
These findings demonstrate the generalization capability of the inpainting pre-training approach, allowing the model to better extract generic representations from the data and improving the final performance.

Clearly, a further key aspect to properly train the DL model in a robust way is the design of the training dataset. Particularly, two aspects need to be considered in case of \mbox{TanDEM-X} InSAR data: (i) to correctly fit the DL model with all possible different acquisition geometries and (ii) to have a balanced training dataset over the different land cover classes.
In our study, we rely on the use of the $h_{amb}$ as representative of the multiple acquisition geometries possible with the \mbox{TanDEM-X} InSAR system.
In our experiments over Pennsylvania test region, we observe that a key aspect when using only a small amount of the labeled data, is the proper selection of a subset of representative \mbox{TanDEM-X} images acquired with different $h_{amb}$, 
since using only images acquired with very similar geometries in the DST leads to unreliable forest classification.
Such an effect is also seen over the Amazon rainforest, where no \mbox{TanDEM-X} images acquired with $h_{amb} < 30~\mathrm{m}$ are overlapping the reference patches. Indeed, in the intercomparison with the ESA CCI HRLC map, a worse performance accuracy is achieved with \mbox{TanDEM-X} images acquired with $h_{amb} \in [20~\mathrm{m} - 30~\mathrm{m}$].
Finally, with respect to the balancing of the training dataset, the Pennsylvania landscape with 60\% of the labeled data corresponding to the forest class allows for selecting a well-balanced training dataset for the different conducted experiments. When considering the Amazon rainforest, we had to discard a large number of available reference patches acquired over dense rainforest only, in order to maintain a similar balancing.
\section{Conclusion}
\label{sec:conclusions}

In this study we successfully demonstrated  the effectiveness of deep convolutional neural networks for mapping forests using \mbox{TanDEM-X} bistatic InSAR acquisitions at a very fine spatial resolution of only 6~m.
To address the challenge of limited reference data at such a fine resolution, we investigated different self-supervised pre-training approaches.
Specifically, the use of inpainting
and of sufficient unlabeled data, representing all \mbox{TanDEM-X} acquisition
geometries, achieved a reliable 
performance and stability, which allowed for effectively transferring knowledge to a supervised forest mapping downstream task with few labeled data.
The implementation of the self-supervised pre-training strategy is particularly interesting in regions like the Amazon rainforest, where reference labeled data at high resolution is scarce and challenging to obtain. 
By using very few labeled samples, our deep learning model based on self-supervised pre-training is able to radically improve the forest classification performance. This allows for the reliable detection of narrow paths and small clear-cuts in dense forested areas using a single bistatic \mbox{TanDEM-X} acquisition at only 6 m independent pixel spacing.
In conclusion, the contributions of this research offer valuable insights into the field of forest mapping with spaceborne bistatic InSAR data. The combination of deep convolutional neural networks, the proper consideration of the different sensor acquisition geometries and the self-supervised pre-training strategy represent a powerful tool to address the challenges posed by limited referenced data in forest mapping applications at high resolution.
Finally, the proposed method represents a solid and promising starting point for setting up a reliable framework for the generation of large-scale very high-resolution forest maps, especially over tropical forests, based on \mbox{TanDEM-X} InSAR acquisitions.


\section*{Acknowledgments}
The authors would like to thank Prof. R. Dubayah and his
team at the University of Maryland for providing the Pennsylvania forest map.


\setcounter{table}{0}
\appendix
\section{Complete testing results}
\label{appendix}

In this section, we present the complete performance metrics over Pennsylvania's testing dataset and for the different DL approaches.
\mbox{Table~\ref{tab:baseline_metrics}} shows the performance metrics obtained for the \textbf{baseline} case and for the different $h_{\mathrm{amb}}$ test subsets defined in \mbox{Section~\ref{sseq:val_test_strategy}}.
\mbox{Table~\ref{tab:metrics_2013}} shows the performance metrics for the investigated DL approaches using the test subset of \mbox{TanDEM-X} images acquired in descending orbit during 2013.
\mbox{Tables~\ref{tab:metrics_22_low_hoa} - \ref{tab:metrics_02_low_hoa}} depict the performance metrics for the different investigated DL approaches and the different $h_{\mathrm{amb}}$ test subsets when using 22\%, 8\%, and 1.5\% of the labeled data.

\begin{table}
\centering
    \caption{Performance metrics for the baseline DL approach (using 100\% of the labeled data) and for the different $h_{\mathrm{amb}}$ test subsets defined in \mbox{Section~\ref{sseq:val_test_strategy}.}}
    \label{tab:baseline_metrics}
\resizebox{.9\columnwidth}{!}{
\begin{tabular}{cccccccccc}\hline
\multicolumn{4}{c}{\textbf{Test subset: Short $h_{\mathrm{amb}}$}} & \multicolumn{3}{c}{Forest} & \multicolumn{3}{c}{Non-forest} \\ 
\multicolumn{1}{c}{DL approach} & Run Nr. & OA & F$^w_1$-score & Precision & Recall & F$_1$-score & Precision & Recall & F$_1$-score \\ \hline
\multirow{4}{*}{\textbf{Baseline}} & 1 & 0.9199 & 0.9194 & 0.9208 & 0.9519 & 0.9361 & 0.9184 & 0.8685 & 0.8927 \\
 & 2 & 0.9210 & 0.9208 & 0.9272 & 0.9462 & 0.9366 & 0.9107 & 0.8806 & 0.8954 \\
 & 3 & 0.9208 & 0.9205 & 0.9258 & 0.9473 & 0.9364 & 0.9121 & 0.8782 & 0.8948 \\
 & Mean & \textbf{0.9206} & \textbf{0.9202} & \textbf{0.9246} & \textbf{0.9485} & \textbf{0.9364} & \textbf{0.9137} & \textbf{0.8758} & \textbf{0.8943} \\ \hline
\multicolumn{4}{c}{\textbf{Test subset: Mid $h_{\mathrm{amb}}$}} & \multicolumn{3}{c}{Forest} & \multicolumn{3}{c}{Non-forest} \\
\multicolumn{1}{c}{DL approach} & Run Nr. & OA & F$^w_1$-score & Precision & Recall & F$_1$-score & Precision & Recall & F$_1$-score \\ \hline
\multirow{4}{*}{\textbf{Baseline}} & 1 & 0.9097 & 0.9093 & 0.9161 & 0.9379 & 0.9269 & 0.8989 & 0.8654 & 0.8818 \\
 & 2 & 0.9106 & 0.9105 & 0.9235 & 0.9306 & 0.9270 & 0.8899 & 0.8792 & 0.8845 \\
 & 3 & 0.9106 & 0.9104 & 0.9192 & 0.9359 & 0.9275 & 0.8966 & 0.8711 & 0.8837 \\
 & Mean & \textbf{0.9103} & \textbf{0.9101} & \textbf{0.9196} & \textbf{0.9348} & \textbf{0.9271} & \textbf{0.8952} & \textbf{0.8719} & \textbf{0.8833} \\ \hline
\multicolumn{4}{c}{\textbf{Test subset: Large $h_{\mathrm{amb}}$}} & \multicolumn{3}{c}{Forest} & \multicolumn{3}{c}{Non-forest} \\
\multicolumn{1}{l}{DL approach} & Run Nr. & OA & F$^w_1$-score & Precision & Recall & F$_1$-score & Precision & Recall & F$_1$-score \\ \hline
\multirow{4}{*}{\textbf{Baseline}} & 1 & 0.9197 & 0.9196 & 0.9318 & 0.9376 & 0.9347 & 0.9002 & 0.8914 & 0.8958 \\
 & 2 & 0.9190 & 0.9187 & 0.9248 & 0.9446 & 0.9346 & 0.9093 & 0.8784 & 0.8936 \\
 & 3 & 0.9190 & 0.9188 & 0.9283 & 0.9405 & 0.9343 & 0.9038 & 0.8851 & 0.8943 \\
 & Mean & \textbf{0.9192} & \textbf{0.9191} & \textbf{0.9283} & \textbf{0.9409} & \textbf{0.9345} & \textbf{0.9045} & \textbf{0.8850} & \textbf{0.8946} \\ \hline
\end{tabular}
}
\end{table}

\begin{table}
\centering
    \caption{Performance metrics for the investigated DL approaches using the test subset of \mbox{TanDEM-X} images acquired in descending orbit during 2013.}
    \label{tab:metrics_2013}
\resizebox{.9\columnwidth}{!}{
\begin{tabular}{cccccccccc}\hline
\multicolumn{4}{c}{\textbf{Test subset: Large $h_{\mathrm{amb}}$}} & \multicolumn{3}{c}{Forest} & \multicolumn{3}{c}{Non-forest} \\
\multicolumn{1}{l}{DL approach} & Run Nr. & OA & F$^w_1$-score & Precision & Recall & F$_1$-score & Precision & Recall & F$_1$-score \\ \hline
\multirow{4}{*}{\textbf{Baseline}} & 1 & 0.8441 & 0.8443 & 0.8752 & 0.8627 & 0.8689 & 0.7992 & 0.8163 & 0.8077 \\
 & 2 & 0.8328 & 0.8324 & 0.8535 & 0.8703 & 0.8618 & 0.8004 & 0.7768 & 0.7884 \\
 & 3 & 0.8338 & 0.8429 & 0.8712 & 0.8597 & 0.8654 & 0.8092 & 0.7968 & 0.8030 \\
 & Mean & \textbf{0.8369} & \textbf{0.8399} & \textbf{0.8666} & \textbf{0.8642} & \textbf{0.8654} & \textbf{0.8029} & \textbf{0.7966} & \textbf{0.7997} \\
\multirow{4}{*}{\textbf{FSL}} & 1 & 0.8171 & 0.8172 & 0.8489 & 0.8450 & 0.8469 & 0.7700 & 0.7754 & 0.7727 \\
 & 2 & 0.8323 & 0.8325 & 0.8630 & 0.8560 & 0.8595 & 0.7875 & 0.7970 & 0.7922 \\
 & 3 & 0.8328 & 0.8332 & 0.8669 & 0.8516 & 0.8592 & 0.7840 & 0.8048 & 0.7943 \\
 & Mean & \textbf{0.8274} & \textbf{0.8276} & \textbf{0.8596} & \textbf{0.8509} & \textbf{0.8552} & \textbf{0.7805} & \textbf{0.7924} & \textbf{0.7864} \\
\multirow{4}{*}{\textbf{SSL-In E+D}} & 1 & 0.8356 & 0.8357 & 0.8645 & 0.8603 & 0.8624 & 0.7929 & 0.7985 & 0.7957 \\
 & 2 & 0.8361 & 0.8353 & 0.8507 & 0.8809 & 0.8655 & 0.8121 & 0.7691 & 0.7900 \\
 & 3 & 0.8359 & 0.8348 & 0.8473 & 0.8856 & 0.8661 & 0.8168 & 0.7616 & 0.7882 \\
 & Mean & \textbf{0.8359} & \textbf{0.8353} & \textbf{0.8542} & \textbf{0.8756} & \textbf{0.8647} & \textbf{0.8073} & \textbf{0.7764} & \textbf{0.7913} \\
\multirow{4}{*}{\textbf{SSL-In D}} & 1 & 0.8360 & 0.8353 & 0.8514 & 0.8797 & 0.8653 & 0.8110 & 0.7707 & 0.7903 \\
 & 2 & 0.8284 & 0.8291 & 0.8706 & 0.8381 & 0.8541 & 0.7710 & 0.8139 & 0.7919 \\
 & 3 & 0.8314 & 0.8300 & 0.8407 & 0.8864 & 0.8630 & 0.8154 & 0.7491 & 0.7808 \\
 & Mean & \textbf{0.8319} & \textbf{0.8315} & \textbf{0.8542} & \textbf{0.8681} & \textbf{0.8608} & \textbf{0.7991} & \textbf{0.7779} & \textbf{0.7877}
\\ \hline
\end{tabular}
}
\end{table}

\begin{table*}[]
\centering
    \caption{Performance metrics, using 22\% of the labeled data, for the different DL approaches and test subsets. 
    }
    \label{tab:metrics_22_low_hoa}
\resizebox{.5\columnwidth}{!}{
\begin{tabular}{cccccccccc}\hline
\multicolumn{4}{c}{\textbf{Test subset: Short $h_{\mathrm{amb}}$}} & \multicolumn{3}{c}{Forest} & \multicolumn{3}{c}{Non-forest} \\ 
\multicolumn{1}{c}{DL approach} & Run Nr. & OA & F$^w_1$-score & Precision & Recall & F$_1$-score & Precision & Recall & F$_1$-score \\ \hline
\multirow{4}{*}{\textbf{FSL}} & 1 & 0.9153 & 0.9154 & 0.9330 & 0.9293 & 0.9312 & 0.8872 & 0.8929 & 0.8901 \\
 & 2 & 0.9163 & 0.9163 & 0.9324 & 0.9317 & 0.9321 & 0.8905 & 0.8916 & 0.8910 \\
 & 3 & 0.9149 & 0.9150 & 0.9328 & 0.9288 & 0.9308 & 0.8864 & 0.8926 & 0.8895 \\
 & Mean & \textbf{0.9155} & \textbf{0.9156} & \textbf{0.9327} & \textbf{0.9299} & \textbf{0.9314} & \textbf{0.8880} & \textbf{0.8924} & \textbf{0.8902} \\
\multirow{4}{*}{\textbf{SSL-Id E+D}} & 1 & 0.9063 & 0.9063 & 0.9254 & 0.9223 & 0.9238 & 0.8759 & 0.8806 & 0.8782 \\
 & 2 & 0.9082 & 0.9084 & 0.9333 & 0.9165 & 0.9248 & 0.8697 & 0.8948 & 0.8821 \\
 & 3 & 0.9118 & 0.9117 & 0.9244 & 0.9333 & 0.9288 & 0.8912 & 0.8774 & 0.8842 \\
 & Mean & \textbf{0.9088} & \textbf{0.9088} & \textbf{0.9277} & \textbf{0.9240} & \textbf{0.9258} & \textbf{0.8789} & \textbf{0.8843} & \textbf{0.8815} \\
\multirow{4}{*}{\textbf{SSL-Id D}} & 1 & 0.8899 & 0.8900 & 0.9149 & 0.9055 & 0.9102 & 0.8507 & 0.8647 & 0.8577 \\
 & 2 & 0.8982 & 0.8982 & 0.9171 & 0.9178 & 0.9175 & 0.8679 & 0.8667 & 0.8673 \\
 & 3 & 0.8874 & 0.8879 & 0.9244 & 0.8900 & 0.9069 & 0.8334 & 0.8831 & 0.8575 \\
 & Mean & \textbf{0.8918} & \textbf{0.8920} & \textbf{0.9188} & \textbf{0.9044} & \textbf{0.9115} & \textbf{0.8507} & \textbf{0.8715} & \textbf{0.8608} \\
\multirow{4}{*}{\textbf{SSL-In E+D}} & 1 & 0.9076 & 0.9078 & 0.9294 & 0.9200 & 0.9247 & 0.8736 & 0.8877 & 0.8806 \\
 & 2 & 0.9160 & 0.9161 & 0.9370 & 0.9258 & 0.9314 & 0.8832 & 0.9001 & 0.8916 \\
 & 3 & 0.9167 & 0.9167 & 0.9309 & 0.9341 & 0.9325 & 0.8936 & 0.8888 & 0.8912 \\
 & Mean & \textbf{0.9134} & \textbf{0.9135} & \textbf{0.9324} & \textbf{0.9266} & \textbf{0.9295} & \textbf{0.8835} & \textbf{0.8922} & \textbf{0.8878} \\
\multirow{4}{*}{\textbf{SSL-In D}} & 1 & 0.9071 & 0.9074 & 0.9352 & 0.9124 & 0.9237 & 0.8646 & 0.8985 & 0.8812 \\
 & 2 & 0.9094 & 0.9093 & 0.9221 & 0.9317 & 0.9269 & 0.8885 & 0.8736 & 0.8810 \\
 & 3 & 0.9083 & 0.9083 & 0.9249 & 0.9264 & 0.9257 & 0.8816 & 0.8792 & 0.8804 \\
 & Mean & \textbf{0.9083} & \textbf{0.9083} & \textbf{0.9274} & \textbf{0.9235} & \textbf{0.9254} & \textbf{0.8782} & \textbf{0.8838} & \textbf{0.8809} \\ \hline
\multicolumn{4}{c}{\textbf{Test subset: Mid $h_{\mathrm{amb}}$}} & \multicolumn{3}{c}{Forest} & \multicolumn{3}{c}{Non-forest} \\
\multicolumn{1}{c}{DL approach} & Run Nr. & OA & F$^w_1$-score & Precision & Recall & F$_1$-score & Precision & Recall & F$_1$-score \\ \hline
\multirow{4}{*}{\textbf{FSL}} & 1 & 0.9007 & 0.9005 & 0.9129 & 0.9256 & 0.9192 & 0.8809 & 0.8616 & 0.8711 \\
 & 2 & 0.8983 & 0.8978 & 0.9035 & 0.9331 & 0.9181 & 0.8895 & 0.8439 & 0.8661 \\
 & 3 & 0.8967 & 0.8965 & 0.9091 & 0.9232 & 0.9161 & 0.8766 & 0.8554 & 0.8658 \\
 & Mean & \textbf{0.8986} & \textbf{0.8983} & \textbf{0.9085} & \textbf{0.9273} & \textbf{0.9178} & \textbf{0.8823} & \textbf{0.8536} & \textbf{0.8677} \\
\multirow{4}{*}{\textbf{SSL-Id E+D}} & 1 & 0.8900 & 0.8893 & 0.8947 & 0.9292 & 0.9116 & 0.8819 & 0.8286 & 0.8544 \\
 & 2 & 0.8908 & 0.8903 & 0.9000 & 0.9237 & 0.9117 & 0.8753 & 0.8392 & 0.8569 \\
 & 3 & 0.8913 & 0.8906 & 0.8946 & 0.9318 & 0.9128 & 0.8856 & 0.8279 & 0.8558 \\
 & Mean & \textbf{0.8907} & \textbf{0.8901} & \textbf{0.8964} & \textbf{0.9282} & \textbf{0.9120} & \textbf{0.8809} & \textbf{0.8319} & \textbf{0.8557} \\
\multirow{4}{*}{\textbf{SSL-Id D}} & 1 & 0.8750 & 0.8739 & 0.8761 & 0.9263 & 0.9005 & 0.8732 & 0.7947 & 0.8321 \\
 & 2 & 0.8743 & 0.8732 & 0.8766 & 0.9243 & 0.8998 & 0.8704 & 0.7960 & 0.8315 \\
 & 3 & 0.8782 & 0.8774 & 0.8842 & 0.9212 & 0.9023 & 0.8678 & 0.8109 & 0.8384 \\
 & Mean & \textbf{0.8758} & \textbf{0.8748} & \textbf{0.8790} & \textbf{0.9239} & \textbf{0.9009} & \textbf{0.8705} & \textbf{0.8005} & \textbf{0.8340} \\
\multirow{4}{*}{\textbf{SSL-In E+D}} & 1 & 0.9013 & 0.9009 & 0.9079 & 0.9330 & 0.9203 & 0.8903 & 0.8516 & 0.8705 \\
 & 2 & 0.9027 & 0.9026 & 0.9179 & 0.9231 & 0.9205 & 0.8784 & 0.8707 & 0.8745 \\
 & 3 & 0.8997 & 0.8990 & 0.8998 & 0.9403 & 0.9196 & 0.8994 & 0.8360 & 0.8665 \\
 & Mean & \textbf{0.9012} & \textbf{0.9008} & \textbf{0.9085} & \textbf{0.9321} & \textbf{0.9201} & \textbf{0.8894} & \textbf{0.8528} & \textbf{0.8705} \\
\multirow{4}{*}{\textbf{SSL-In D}} & 1 & 0.8867 & 0.8862 & 0.8971 & 0.9198 & 0.9083 & 0.8692 & 0.8347 & 0.8516 \\
 & 2 & 0.8861 & 0.8850 & 0.8845 & 0.9355 & 0.9093 & 0.8890 & 0.8086 & 0.8469 \\
 & 3 & 0.8863 & 0.8853 & 0.8879 & 0.9312 & 0.9090 & 0.8833 & 0.8158 & 0.8482 \\
 & Mean & \textbf{0.8864} & \textbf{0.8855} & \textbf{0.8898} & \textbf{0.9288} & \textbf{0.9089} & \textbf{0.8805} & \textbf{0.8197} & \textbf{0.8489} \\ \hline
\multicolumn{4}{c}{\textbf{Test subset: Large $h_{\mathrm{amb}}$}} & \multicolumn{3}{c}{Forest} & \multicolumn{3}{c}{Non-forest} \\
\multicolumn{1}{l}{DL approach} & Run Nr. & OA & F$^w_1$-score & Precision & Recall & F$_1$-score & Precision & Recall & F$_1$-score \\ \hline
\multirow{4}{*}{\textbf{FSL}} & 1 & 0.9136 & 0.9134 & 0.9220 & 0.9384 & 0.9301 & 0.8997 & 0.8744 & 0.8869 \\
 & 2 & 0.9107 & 0.9102 & 0.9140 & 0.9430 & 0.9282 & 0.9050 & 0.8596 & 0.8817 \\
 & 3 & 0.9117 & 0.9112 & 0.9150 & 0.9435 & 0.9290 & 0.9060 & 0.8613 & 0.8831 \\
 & Mean & \textbf{0.9120} & \textbf{0.9116} & \textbf{0.9170} & \textbf{0.9416} & \textbf{0.9291} & \textbf{0.9036} & \textbf{0.8651} & \textbf{0.8839} \\
\multirow{4}{*}{\textbf{SSL-Id E+D}} & 1 & 0.9055 & 0.9049 & 0.9057 & 0.9441 & 0.9245 & 0.9052 & 0.8446 & 0.8738 \\
 & 2 & 0.9055 & 0.9050 & 0.9105 & 0.9380 & 0.9240 & 0.8969 & 0.8540 & 0.8750 \\
 & 3 & 0.9045 & 0.9040 & 0.9090 & 0.9380 & 0.9233 & 0.8967 & 0.8515 & 0.8735 \\
 & Mean & \textbf{0.9052} & \textbf{0.9046} & \textbf{0.9084} & \textbf{0.9400} & \textbf{0.9239} & \textbf{0.8996} & \textbf{0.8500} & \textbf{0.8741} \\
\multirow{4}{*}{\textbf{SSL-Id D}} & 1 & 0.8885 & 0.8868 & 0.8776 & 0.9507 & 0.9127 & 0.9101 & 0.7902 & 0.8459 \\
 & 2 & 0.8877 & 0.8861 & 0.8782 & 0.9483 & 0.9119 & 0.9063 & 0.7919 & 0.8453 \\
 & 3 & 0.8945 & 0.8936 & 0.8933 & 0.9401 & 0.9161 & 0.8966 & 0.8224 & 0.8579 \\
 & Mean & \textbf{0.8902} & \textbf{0.8888} & \textbf{0.8830} & \textbf{0.9464} & \textbf{0.9136} & \textbf{0.9043} & \textbf{0.8015} & \textbf{0.8497} \\
\multirow{4}{*}{\textbf{SSL-In E+D}} & 1 & 0.9077 & 0.9070 & 0.9051 & 0.9489 & 0.9265 & 0.9124 & 0.8427 & 0.8762 \\
 & 2 & 0.9106 & 0.9101 & 0.9128 & 0.9443 & 0.9283 & 0.9067 & 0.8573 & 0.8813 \\
 & 3 & 0.9092 & 0.9085 & 0.9066 & 0.9496 & 0.9276 & 0.9139 & 0.8452 & 0.8782 \\
 & Mean & \textbf{0.9092} & \textbf{0.9085} & \textbf{0.9082} & \textbf{0.9476} & \textbf{0.9275} & \textbf{0.9110} & \textbf{0.8484} & \textbf{0.8786} \\
\multirow{4}{*}{\textbf{SSL-In D}} & 1 & 0.9020 & 0.9013 & 0.9025 & 0.9418 & 0.9217 & 0.9011 & 0.8391 & 0.8690 \\
 & 2 & 0.8978 & 0.8967 & 0.8931 & 0.9465 & 0.9190 & 0.9065 & 0.8207 & 0.8615 \\
 & 3 & 0.8991 & 0.8981 & 0.8946 & 0.9469 & 0.9200 & 0.9075 & 0.8235 & 0.8635 \\
 & Mean & \textbf{0.8996} & \textbf{0.8987} & \textbf{0.8967} & \textbf{0.9451} & \textbf{0.9202} & \textbf{0.9050} & \textbf{0.8278} & \textbf{0.8647} \\ \hline
\end{tabular}
}
\end{table*}

\begin{table}[]
\centering
    \caption{Performance metrics, using 8\% of the labeled data, for the different DL approaches and test subsets.}
        \label{tab:metrics_10_low_hoa}
\resizebox{.5\columnwidth}{!}{
\begin{tabular}{cccccccccc}\hline
\multicolumn{4}{c}{\textbf{Test subset: Short $h_{\mathrm{amb}}$}} & \multicolumn{3}{c}{Forest} & \multicolumn{3}{c}{Non-forest} \\ 
\multicolumn{1}{c}{DL approach} & Run Nr. & OA & F$^w_1$-score & Precision & Recall & F$_1$-score & Precision & Recall & F$_1$-score \\ \hline
\multirow{4}{*}{\textbf{FSL}} & 1 & 0.9085 & 0.9078 & 0.9068 & 0.9491 & 0.9275 & 0.9117 & 0.8433 & 0.8762 \\
 & 2 & 0.9088 & 0.9080 & 0.9057 & 0.9510 & 0.9278 & 0.9145 & 0.8410 & 0.8762 \\
 & 3 & 0.9127 & 0.9126 & 0.9256 & 0.9333 & 0.9294 & 0.8915 & 0.8795 & 0.8855 \\
 & Mean & \textbf{0.9100} & \textbf{0.9095} & \textbf{0.9127} & \textbf{0.9445} & \textbf{0.9282} & \textbf{0.9059} & \textbf{0.8546} & \textbf{0.8793} \\
\multirow{4}{*}{\textbf{SSL-Id E+D}} & 1 & 0.9038 & 0.9034 & 0.9110 & 0.9353 & 0.9230 & 0.8914 & 0.8533 & 0.8719 \\
 & 2 & 0.8996 & 0.8980 & 0.8869 & 0.9594 & 0.9217 & 0.9249 & 0.8036 & 0.8600 \\
 & 3 & 0.9070 & 0.9066 & 0.9139 & 0.9373 & 0.9255 & 0.8951 & 0.8582 & 0.8762 \\
 & Mean & \textbf{0.9035} & \textbf{0.9027} & \textbf{0.9039} & \textbf{0.9440} & \textbf{0.9234} & \textbf{0.9038} & \textbf{0.8384} & \textbf{0.8694} \\
\multirow{4}{*}{\textbf{SSL-Id D}} & 1 & 0.8947 & 0.8937 & 0.8927 & 0.9423 & 0.9169 & 0.8984 & 0.8182 & 0.8564 \\
 & 2 & 0.8975 & 0.8970 & 0.9042 & 0.9325 & 0.9182 & 0.8859 & 0.8414 & 0.8631 \\
 & 3 & 0.8913 & 0.8904 & 0.8928 & 0.9359 & 0.9139 & 0.8885 & 0.8196 & 0.8527 \\
 & Mean & \textbf{0.8945} & \textbf{0.8937} & \textbf{0.8966} & \textbf{0.9369} & \textbf{0.9163} & \textbf{0.8909} & \textbf{0.8264} & \textbf{0.8574} \\
\multirow{4}{*}{\textbf{SSL-In E+D}} & 1 & 0.9144 & 0.9142 & 0.9237 & 0.9386 & 0.9311 & 0.8988 & 0.8756 & 0.8870 \\
 & 2 & 0.9124 & 0.9119 & 0.9141 & 0.9469 & 0.9302 & 0.9095 & 0.8571 & 0.8825 \\
 & 3 & 0.9131 & 0.9130 & 0.9255 & 0.9342 & 0.9299 & 0.8928 & 0.8793 & 0.8860 \\
 & Mean & \textbf{0.9133} & \textbf{0.9130} & \textbf{0.9211} & \textbf{0.9399} & \textbf{0.9304} & \textbf{0.9004} & \textbf{0.8707} & \textbf{0.8852} \\
\multirow{4}{*}{\textbf{SSL-In D}} & 1 & 0.9057 & 0.9056 & 0.9190 & 0.9289 & 0.9239 & 0.8838 & 0.8685 & 0.8761 \\
 & 2 & 0.9032 & 0.9023 & 0.9001 & 0.9482 & 0.9235 & 0.9090 & 0.8310 & 0.8683 \\
 & 3 & 0.9034 & 0.9035 & 0.9225 & 0.9206 & 0.9216 & 0.8729 & 0.8759 & 0.8744 \\
 & Mean & \textbf{0.9041} & \textbf{0.9038} & \textbf{0.9139} & \textbf{0.9326} & \textbf{0.9230} & \textbf{0.8886} & \textbf{0.8585} & \textbf{0.8729} \\ \hline
\multicolumn{4}{c}{\textbf{Test subset: Mid $h_{\mathrm{amb}}$}} & \multicolumn{3}{c}{Forest} & \multicolumn{3}{c}{Non-forest} \\
\multicolumn{1}{c}{DL approach} & Run Nr. & OA & F$^w_1$-score & Precision & Recall & F$_1$-score & Precision & Recall & F$_1$-score \\ \hline
\multirow{4}{*}{\textbf{FSL}} & 1 & 0.875 & 0.8739 & 0.8775 & 0.9243 & 0.9003 & 0.8705 & 0.7978 & 0.8326 \\
 & 2 & 0.8714 & 0.8696 & 0.8652 & 0.9352 & 0.8988 & 0.8836 & 0.7716 & 0.8238 \\
 & 3 & 0.8828 & 0.8823 & 0.8925 & 0.9187 & 0.9054 & 0.8665 & 0.8266 & 0.8461 \\
 & Mean & \textbf{0.8764} & \textbf{0.8753} & \textbf{0.8784} & \textbf{0.9261} & \textbf{0.9015} & \textbf{0.8735} & \textbf{0.7987} & \textbf{0.8342} \\
\multirow{4}{*}{\textbf{SSL-Id E+D}} & 1 & 0.8737 & 0.8721 & 0.8702 & 0.9321 & 0.9001 & 0.8802 & 0.7821 & 0.8283 \\
 & 2 & 0.8624 & 0.8595 & 0.8491 & 0.942 & 0.8931 & 0.8902 & 0.7377 & 0.8068 \\
 & 3 & 0.87 & 0.8689 & 0.8746 & 0.9187 & 0.8961 & 0.8617 & 0.7937 & 0.8263 \\
 & Mean & \textbf{0.8687} & \textbf{0.8668} & \textbf{0.8646} & \textbf{0.9309} & \textbf{0.8964} & \textbf{0.8774} & \textbf{0.7712} & \textbf{0.8205} \\
\multirow{4}{*}{\textbf{SSL-Id D}} & 1 & 0.8552 & 0.8531 & 0.8527 & 0.9221 & 0.886 & 0.86 & 0.7503 & 0.8014 \\
 & 2 & 0.8648 & 0.8634 & 0.8672 & 0.9193 & 0.8925 & 0.8605 & 0.7794 & 0.8179 \\
 & 3 & 0.8498 & 0.8478 & 0.8507 & 0.9145 & 0.8814 & 0.8482 & 0.7485 & 0.7952 \\
 & Mean & \textbf{0.8566} & \textbf{0.8548} & \textbf{0.8569} & \textbf{0.9186} & \textbf{0.8866} & \textbf{0.8562} & \textbf{0.7594} & \textbf{0.8048} \\
\multirow{4}{*}{\textbf{SSL-In E+D}} & 1 & 0.8843 & 0.8838 & 0.894 & 0.9195 & 0.9065 & 0.8679 & 0.8291 & 0.8481 \\
 & 2 & 0.8857 & 0.8846 & 0.8838 & 0.9359 & 0.9091 & 0.8893 & 0.8072 & 0.8462 \\
 & 3 & 0.885 & 0.8842 & 0.888 & 0.9288 & 0.9079 & 0.8798 & 0.8164 & 0.8469 \\
 & Mean & \textbf{0.8850} & \textbf{0.8842} & \textbf{0.8886} & \textbf{0.9281} & \textbf{0.9078} & \textbf{0.8790} & \textbf{0.8176} & \textbf{0.8471} \\
\multirow{4}{*}{\textbf{SSL-In D}} & 1 & 0.8779 & 0.8768 & 0.8798 & 0.9265 & 0.9026 & 0.8744 & 0.8017 & 0.8365 \\
 & 2 & 0.8726 & 0.8707 & 0.8661 & 0.936 & 0.8997 & 0.8852 & 0.7732 & 0.8254 \\
 & 3 & 0.8771 & 0.876 & 0.8786 & 0.9267 & 0.902 & 0.8743 & 0.7993 & 0.8352 \\
 & Mean & \textbf{0.8759} & \textbf{0.8745} & \textbf{0.8748} & \textbf{0.9297} & \textbf{0.9014} & \textbf{0.8780} & \textbf{0.7914} & \textbf{0.8324} \\ \hline
\multicolumn{4}{c}{\textbf{Test subset: Large $h_{\mathrm{amb}}$}} & \multicolumn{3}{c}{Forest} & \multicolumn{3}{c}{Non-forest} \\
\multicolumn{1}{l}{DL approach} & Run Nr. & OA & F$^w_1$-score & Precision & Recall & F$_1$-score & Precision & Recall & F$_1$-score \\ \hline
\multirow{4}{*}{\textbf{FSL}} & 1 & 0.9073 & 0.9069 & 0.9127 & 0.9384 & 0.9254 & 0.898 & 0.858 & 0.8776 \\
 & 2 & 0.9022 & 0.9016 & 0.9048 & 0.9392 & 0.9217 & 0.8977 & 0.8437 & 0.8698 \\
 & 3 & 0.9112 & 0.9111 & 0.9254 & 0.9301 & 0.9277 & 0.8885 & 0.8813 & 0.8849 \\
 & Mean & \textbf{0.9069} & \textbf{0.9065} & \textbf{0.9143} & \textbf{0.9359} & \textbf{0.9249} & \textbf{0.8947} & \textbf{0.8610} & \textbf{0.8774} \\
\multirow{4}{*}{\textbf{SSL-Id E+D}} & 1 & 0.887 & 0.8851 & 0.8736 & 0.9536 & 0.9118 & 0.9141 & 0.7817 & 0.8427 \\
 & 2 & 0.8891 & 0.8874 & 0.8781 & 0.951 & 0.9131 & 0.9107 & 0.7912 & 0.8468 \\
 & 3 & 0.8984 & 0.8982 & 0.9133 & 0.9217 & 0.9174 & 0.8742 & 0.8615 & 0.8678 \\
 & Mean & \textbf{0.8915} & \textbf{0.8902} & \textbf{0.8883} & \textbf{0.9421} & \textbf{0.9141} & \textbf{0.8997} & \textbf{0.8115} & \textbf{0.8524} \\
\multirow{4}{*}{\textbf{SSL-Id D}} & 1 & 0.8901 & 0.8891 & 0.8892 & 0.9375 & 0.9127 & 0.8918 & 0.8151 & 0.8517 \\
 & 2 & 0.8961 & 0.8959 & 0.9104 & 0.9211 & 0.9157 & 0.8728 & 0.8566 & 0.8646 \\
 & 3 & 0.8745 & 0.8719 & 0.8593 & 0.9508 & 0.9027 & 0.9064 & 0.7537 & 0.823 \\
 & Mean & \textbf{0.8869} & \textbf{0.8856} & \textbf{0.8863} & \textbf{0.9365} & \textbf{0.9104} & \textbf{0.8903} & \textbf{0.8085} & \textbf{0.8464} \\
\multirow{4}{*}{\textbf{SSL-In E+D}} & 1 & 0.9136 & 0.9135 & 0.9262 & 0.9334 & 0.9298 & 0.8933 & 0.8824 & 0.8878 \\
 & 2 & 0.9092 & 0.9087 & 0.9107 & 0.9445 & 0.9273 & 0.9067 & 0.8535 & 0.8793 \\
 & 3 & 0.9107 & 0.9103 & 0.9156 & 0.941 & 0.9282 & 0.9024 & 0.8628 & 0.8822 \\
 & Mean & \textbf{0.9112} & \textbf{0.9108} & \textbf{0.9175} & \textbf{0.9396} & \textbf{0.9284} & \textbf{0.9008} & \textbf{0.8662} & \textbf{0.8831} \\
\multirow{4}{*}{\textbf{SSL-In D}} & 1 & 0.9014 & 0.9007 & 0.901 & 0.9427 & 0.9214 & 0.9022 & 0.8361 & 0.8679 \\
 & 2 & 0.9008 & 0.8999 & 0.8985 & 0.9448 & 0.921 & 0.9048 & 0.8311 & 0.8664 \\
 & 3 & 0.9042 & 0.9038 & 0.9121 & 0.9336 & 0.9227 & 0.8908 & 0.8576 & 0.8739 \\
 & Mean & \textbf{0.9021} & \textbf{0.9015} & \textbf{0.9039} & \textbf{0.9404} & \textbf{0.9217} & \textbf{0.8993} & \textbf{0.8416} & \textbf{0.8694} \\ \hline
\end{tabular}
}
\end{table}

\begin{table*}[]
\centering
    \caption{Performance metrics, using 1.5\% of the labeled data, for the different DL approaches and test subsets.}
        \label{tab:metrics_02_low_hoa}
\resizebox{.5\columnwidth}{!}{
\begin{tabular}{cccccccccc}\hline
\multicolumn{4}{c}{\textbf{Test subset: Short $h_{\mathrm{amb}}$}} & \multicolumn{3}{c}{Forest} & \multicolumn{3}{c}{Non-forest} \\ 
\multicolumn{1}{c}{DL approach} & Run Nr. & OA & F$^w_1$-score & Precision & Recall & F$_1$-score & Precision & Recall & F$_1$-score \\ \hline
\multirow{4}{*}{\textbf{FSL}} & 1 & 0.8929 & 0.8927 & 0.9085 & 0.9187 & 0.9136 & 0.8671 & 0.8515 & 0.8592 \\
 & 2 & 0.8970 & 0.8970 & 0.9174 & 0.9153 & 0.9163 & 0.8644 & 0.8677 & 0.8661 \\
 & 3 & 0.8979 & 0.8975 & 0.9058 & 0.9312 & 0.9183 & 0.8843 & 0.8446 & 0.8640 \\
 & Mean & \textbf{0.8959} & \textbf{0.8957} & \textbf{0.9106} & \textbf{0.9217} & \textbf{0.9161} & \textbf{0.8719} & \textbf{0.8546} & \textbf{0.8631} \\
\multirow{4}{*}{\textbf{SSL-Id E+D}} & 1 & 0.8920 & 0.8911 & 0.8940 & 0.9358 & 0.9144 & 0.8885 & 0.8218 & 0.8538 \\
 & 2 & 0.8875 & 0.8874 & 0.9055 & 0.9127 & 0.9091 & 0.8580 & 0.8470 & 0.8525 \\
 & 3 & 0.8802 & 0.8809 & 0.9242 & 0.8775 & 0.9002 & 0.8181 & 0.8844 & 0.8500 \\
 & Mean & \textbf{0.8866} & \textbf{0.8865} & \textbf{0.9079} & \textbf{0.9087} & \textbf{0.9079} & \textbf{0.8549} & \textbf{0.8511} & \textbf{0.8521} \\
\multirow{4}{*}{\textbf{SSL-Id D}} & 1 & 0.8662 & 0.8662 & 0.8925 & 0.8900 & 0.8913 & 0.8242 & 0.8280 & 0.8261 \\
 & 2 & 0.8726 & 0.8727 & 0.8983 & 0.8946 & 0.8964 & 0.8319 & 0.8373 & 0.8346 \\
 & 3 & 0.8638 & 0.8645 & 0.9066 & 0.8684 & 0.8871 & 0.8021 & 0.8564 & 0.8283 \\
 & Mean & \textbf{0.8675} & \textbf{0.8678} & \textbf{0.8991} & \textbf{0.8843} & \textbf{0.8916} & \textbf{0.8194} & \textbf{0.8406} & \textbf{0.8297} \\
\multicolumn{1}{l}{\multirow{4}{*}{\textbf{SSL-In E+D}}} & 1 & 0.9096 & 0.9095 & 0.9230 & 0.9308 & 0.9269 & 0.8874 & 0.8754 & 0.8814 \\
\multicolumn{1}{l}{} & 2 & 0.9058 & 0.9050 & 0.9040 & 0.9478 & 0.9254 & 0.9092 & 0.8384 & 0.8724 \\
\multicolumn{1}{l}{} & 3 & 0.9056 & 0.9051 & 0.9095 & 0.9405 & 0.9247 & 0.8989 & 0.8497 & 0.8736 \\
\multicolumn{1}{l}{} & Mean & \textbf{0.9070} & \textbf{0.9065} & \textbf{0.9122} & \textbf{0.9397} & \textbf{0.9257} & \textbf{0.8985} & \textbf{0.8545} & \textbf{0.8758} \\
\multirow{4}{*}{\textbf{SSL-In D}} & 1 & 0.8983 & 0.8975 & 0.8979 & 0.9422 & 0.9195 & 0.8992 & 0.8280 & 0.8621 \\
 & 2 & 0.8983 & 0.8976 & 0.9011 & 0.9379 & 0.9191 & 0.8933 & 0.8347 & 0.8630 \\
 & 3 & 0.8988 & 0.8979 & 0.8973 & 0.9438 & 0.9200 & 0.9016 & 0.8266 & 0.8625 \\
 & Mean & \textbf{0.8985} & \textbf{0.8977} & \textbf{0.8988} & \textbf{0.9413} & \textbf{0.9195} & \textbf{0.8980} & \textbf{0.8298} & \textbf{0.8625}
\\ \hline
\multicolumn{4}{c}{\textbf{Test subset: Mid $h_{\mathrm{amb}}$}} & \multicolumn{3}{c}{Forest} & \multicolumn{3}{c}{Non-forest} \\
\multicolumn{1}{c}{DL approach} & Run Nr. & OA & F$^w_1$-score & Precision & Recall & F$_1$-score & Precision & Recall & F$_1$-score \\ \hline
\multirow{4}{*}{\textbf{FSL}} & 1 & 0.8527 & 0.8533 & 0.8921 & 0.8631 & 0.8774 & 0.7959 & 0.8365 & 0.8157 \\
 & 2 & 0.8526 & 0.8533 & 0.8946 & 0.8598 & 0.8769 & 0.7930 & 0.8412 & 0.8164 \\
 & 3 & 0.8666 & 0.8662 & 0.8834 & 0.9002 & 0.8917 & 0.8388 & 0.8139 & 0.8262 \\
 & Mean & \textbf{0.8573} & \textbf{0.8576} & \textbf{0.8900} & \textbf{0.8744} & \textbf{0.8820} & \textbf{0.8092} & \textbf{0.8305} & \textbf{0.8194} \\
\multirow{4}{*}{\textbf{SSL-Id E+D}} & 1 & 0.8622 & 0.8608 & 0.8647 & 0.9179 & 0.8905 & 0.8576 & 0.7749 & 0.8142 \\
 & 2 & 0.8595 & 0.8590 & 0.8765 & 0.8961 & 0.8862 & 0.8313 & 0.8022 & 0.8164 \\
 & 3 & 0.8521 & 0.8521 & 0.8793 & 0.8783 & 0.8788 & 0.8096 & 0.8111 & 0.8104 \\
 & Mean & \textbf{0.8579} & \textbf{0.8573} & \textbf{0.8735} & \textbf{0.8974} & \textbf{0.8852} & \textbf{0.8328} & \textbf{0.7961} & \textbf{0.8137} \\
\multirow{4}{*}{\textbf{SSL-Id D}} & 1 & 0.8432 & 0.8424 & 0.8597 & 0.8880 & 0.8736 & 0.8149 & 0.7729 & 0.7934 \\
 & 2 & 0.8472 & 0.8462 & 0.8601 & 0.8953 & 0.8774 & 0.8247 & 0.7718 & 0.7974 \\
 & 3 & 0.8404 & 0.8405 & 0.8715 & 0.8662 & 0.8688 & 0.7923 & 0.7999 & 0.7961 \\
 & Mean & \textbf{0.8436} & \textbf{0.8430} & \textbf{0.8638} & \textbf{0.8832} & \textbf{0.8733} & \textbf{0.8106} & \textbf{0.7815} & \textbf{0.7956} \\
\multirow{4}{*}{\textbf{SSL-In E+D}} & 1 & 0.8792 & 0.8788 & 0.8923 & 0.9122 & 0.9022 & 0.8575 & 0.8275 & 0.8422 \\
 & 2 & 0.8787 & 0.8775 & 0.8781 & 0.9306 & 0.9036 & 0.8800 & 0.7975 & 0.8367 \\
 & 3 & 0.8778 & 0.8766 & 0.8776 & 0.9296 & 0.9028 & 0.8783 & 0.7968 & 0.8356 \\
 & Mean & \textbf{0.8786} & \textbf{0.8776} & \textbf{0.8827} & \textbf{0.9241} & \textbf{0.9029} & \textbf{0.8719} & \textbf{0.8073} & \textbf{0.8382} \\
\multirow{4}{*}{\textbf{SSL-In D}} & 1 & 0.8605 & 0.8585 & 0.8560 & 0.9276 & 0.8904 & 0.8694 & 0.7555 & 0.8085 \\
 & 2 & 0.8612 & 0.8598 & 0.8646 & 0.9160 & 0.8896 & 0.8548 & 0.7752 & 0.8131 \\
 & 3 & 0.8617 & 0.8596 & 0.8572 & 0.9281 & 0.8912 & 0.8705 & 0.7577 & 0.8102 \\
 & Mean & \textbf{0.8611} & \textbf{0.8593} & \textbf{0.8593} & \textbf{0.9239} & \textbf{0.8904} & \textbf{0.8649} & \textbf{0.7628} & \textbf{0.8106}
\\ \hline
\multicolumn{4}{c}{\textbf{Test subset: Large $h_{\mathrm{amb}}$}} & \multicolumn{3}{c}{Forest} & \multicolumn{3}{c}{Non-forest} \\
\multicolumn{1}{l}{DL approach} & Run Nr. & OA & F$^w_1$-score & Precision & Recall & F$_1$-score & Precision & Recall & F$_1$-score \\ \hline
\multirow{4}{*}{\textbf{FSL}} & 1 & 0.8494 & 0.8441 & 0.8251 & 0.9572 & 0.8862 & 0.9092 & 0.6790 & 0.7775 \\
 & 2 & 0.8644 & 0.8617 & 0.8537 & 0.9397 & 0.8946 & 0.8865 & 0.7452 & 0.8097 \\
 & 3 & 0.8776 & 0.8761 & 0.8747 & 0.9340 & 0.9034 & 0.8831 & 0.7883 & 0.8330 \\
 & Mean & \textbf{0.8638} & \textbf{0.8606} & \textbf{0.8512} & \textbf{0.9436} & \textbf{0.8947} & \textbf{0.8929} & \textbf{0.7375} & \textbf{0.8067} \\
\multirow{4}{*}{\textbf{SSL-Id E+D}} & 1 & 0.8485 & 0.8424 & 0.8204 & 0.9637 & 0.8863 & 0.9207 & 0.6661 & 0.7730 \\
 & 2 & 0.8523 & 0.8516 & 0.8692 & 0.8933 & 0.8811 & 0.8234 & 0.7874 & 0.8050 \\
 & 3 & 0.8640 & 0.8618 & 0.8581 & 0.9321 & 0.8936 & 0.8757 & 0.7561 & 0.8115 \\
 & Mean & \textbf{0.8549} & \textbf{0.8519} & \textbf{0.8492} & \textbf{0.9297} & \textbf{0.8870} & \textbf{0.8733} & \textbf{0.7365} & \textbf{0.7965} \\
\multirow{4}{*}{\textbf{SSL-Id D}} & 1 & 0.8477 & 0.8480 & 0.8828 & 0.8664 & 0.8745 & 0.7947 & 0.8180 & 0.8062 \\
 & 2 & 0.8617 & 0.8606 & 0.8684 & 0.9126 & 0.8900 & 0.8496 & 0.7813 & 0.8140 \\
 & 3 & 0.8487 & 0.8485 & 0.8721 & 0.8826 & 0.8773 & 0.8107 & 0.7952 & 0.8028 \\
 & Mean & \textbf{0.8527} & \textbf{0.8524} & \textbf{0.8744} & \textbf{0.8872} & \textbf{0.8806} & \textbf{0.8183} & \textbf{0.7982} & \textbf{0.8077} \\
\multirow{4}{*}{\textbf{SSL-In E+D}} & 1 & 0.8860 & 0.8843 & 0.8762 & 0.9480 & 0.9107 & 0.9054 & 0.7881 & 0.8427 \\
 & 2 & 0.8844 & 0.8827 & 0.8753 & 0.9461 & 0.9093 & 0.9023 & 0.7867 & 0.8405 \\
 & 3 & 0.8718 & 0.8685 & 0.8511 & 0.9585 & 0.9016 & 0.9179 & 0.7346 & 0.8161 \\
 & Mean & \textbf{0.8807} & \textbf{0.8785} & \textbf{0.8675} & \textbf{0.9509} & \textbf{0.9072} & \textbf{0.9085} & \textbf{0.7698} & \textbf{0.8331} \\
\multirow{4}{*}{\textbf{SSL-In D}} & 1 & 0.8674 & 0.8645 & 0.8519 & 0.9486 & 0.8976 & 0.9008 & 0.7391 & 0.8120 \\
 & 2 & 0.8742 & 0.8732 & 0.8798 & 0.9205 & 0.8997 & 0.8643 & 0.8010 & 0.8314 \\
 & 3 & 0.8564 & 0.8521 & 0.8353 & 0.9537 & 0.8906 & 0.9056 & 0.7025 & 0.7912 \\
 & Mean & \textbf{0.8660} & \textbf{0.8633} & \textbf{0.8557} & \textbf{0.9409} & \textbf{0.8960} & \textbf{0.8902} & \textbf{0.7475} & \textbf{0.8115}
\\ \hline
\end{tabular}
}
\end{table*}

\clearpage

\begin{thebibliography}{48}
\expandafter\ifx\csname natexlab\endcsname\relax\def\natexlab#1{#1}\fi
\providecommand{\url}[1]{\texttt{#1}}
\providecommand{\href}[2]{#2}
\providecommand{\path}[1]{#1}
\providecommand{\DOIprefix}{doi:}
\providecommand{\ArXivprefix}{arXiv:}
\providecommand{\URLprefix}{URL: }
\providecommand{\Pubmedprefix}{pmid:}
\providecommand{\doi}[1]{\href{http://dx.doi.org/#1}{\path{#1}}}
\providecommand{\Pubmed}[1]{\href{pmid:#1}{\path{#1}}}
\providecommand{\bibinfo}[2]{#2}
\ifx\xfnm\relax \def\xfnm[#1]{\unskip,\space#1}\fi
\bibitem[{Bruzzone et~al.(2024)Bruzzone, Bovolo, Amodio, Brovelli, Corsi, Defourny, Domingo, Gamba, Kolitzus, Lamarche, Moser, Ottlé, Perantoni, Pesquer and Zanetti}]{esa_hrlc_cci_2024}
\bibinfo{author}{Bruzzone, L.}, \bibinfo{author}{Bovolo, F.}, \bibinfo{author}{Amodio, A.}, \bibinfo{author}{Brovelli, M.}, \bibinfo{author}{Corsi, M.}, \bibinfo{author}{Defourny, P.}, \bibinfo{author}{Domingo, C.}, \bibinfo{author}{Gamba, P.}, \bibinfo{author}{Kolitzus, D.}, \bibinfo{author}{Lamarche, C.}, \bibinfo{author}{Moser, G.}, \bibinfo{author}{Ottlé, C.}, \bibinfo{author}{Perantoni, G.}, \bibinfo{author}{Pesquer, L.}, \bibinfo{author}{Zanetti, M.}, \bibinfo{year}{2024}.
\newblock \bibinfo{title}{{ESA High Resolution Land Cover Climate Change Initiative: High Resolution Land Cover Maps in Amazonia (Eastern Amazonas region) at 10m spatial resolution for 2019 in Geotiff format, v1.2.}}
\newblock \DOIprefix\doi{10.5285/0bc7042123984c69aa45cb6788bfdaa0}.
\bibitem[{Bueso-Bello et~al.(2022)Bueso-Bello, Carcereri, Martone, Gonzalez, Posovszky and Rizzoli}]{buesobello_fnf_2022}
\bibinfo{author}{Bueso-Bello, J.L.}, \bibinfo{author}{Carcereri, D.}, \bibinfo{author}{Martone, M.}, \bibinfo{author}{Gonzalez, C.}, \bibinfo{author}{Posovszky, P.}, \bibinfo{author}{Rizzoli, P.}, \bibinfo{year}{2022}.
\newblock \bibinfo{title}{Deep learning for mapping tropical forests with {TanDEM-X} bistatic {InSAR} data}.
\newblock \bibinfo{journal}{Remote Sensing} \bibinfo{volume}{14}.
\bibitem[{Carcereri et~al.(2024)Carcereri, Rizzoli, Dell’Amore, Bueso-Bello, Ienco and Bruzzone}]{CARCERERI2024114270}
\bibinfo{author}{Carcereri, D.}, \bibinfo{author}{Rizzoli, P.}, \bibinfo{author}{Dell’Amore, L.}, \bibinfo{author}{Bueso-Bello, J.L.}, \bibinfo{author}{Ienco, D.}, \bibinfo{author}{Bruzzone, L.}, \bibinfo{year}{2024}.
\newblock \bibinfo{title}{{Generation of country-scale canopy height maps over Gabon using deep learning and TanDEM-X InSAR data}}.
\newblock \bibinfo{journal}{Remote Sensing of Environment} \bibinfo{volume}{311}, \bibinfo{pages}{114270}.
\newblock \URLprefix \url{https://www.sciencedirect.com/science/article/pii/S0034425724002888}, \DOIprefix\doi{https://doi.org/10.1016/j.rse.2024.114270}.
\bibitem[{Carcereri et~al.(2023)Carcereri, Rizzoli, Ienco and Bruzzone}]{Carcereri2023}
\bibinfo{author}{Carcereri, D.}, \bibinfo{author}{Rizzoli, P.}, \bibinfo{author}{Ienco, D.}, \bibinfo{author}{Bruzzone, L.}, \bibinfo{year}{2023}.
\newblock \bibinfo{title}{A deep learning framework for the estimation of forest height from bistatic {TanDEM-X} data}.
\newblock \bibinfo{journal}{IEEE Journal of Selected Topics in Applied Earth Observations and Remote Sensing} \bibinfo{volume}{16}, \bibinfo{pages}{8334--8352}.
\newblock \DOIprefix\doi{10.1109/JSTARS.2023.3310209}.
\bibitem[{Chazdon et~al.(2016)Chazdon, Brancalion, Laestadius, Bennett-Curry, Buckingham, Kumar, Moll-Rocek, Vieira and Wilson}]{isForest_Chazdon_2016}
\bibinfo{author}{Chazdon, R.L.}, \bibinfo{author}{Brancalion, P.H.}, \bibinfo{author}{Laestadius, L.}, \bibinfo{author}{Bennett-Curry, A.}, \bibinfo{author}{Buckingham, K.}, \bibinfo{author}{Kumar, C.}, \bibinfo{author}{Moll-Rocek, J.}, \bibinfo{author}{Vieira, I.}, \bibinfo{author}{Wilson, S.}, \bibinfo{year}{2016}.
\newblock \bibinfo{title}{{When is a forest a forest? Forest concepts and definitions in the era of forest and landscape restoration}}.
\newblock \bibinfo{journal}{Ambio} \bibinfo{volume}{45}, \bibinfo{pages}{538--550}.
\newblock \DOIprefix\doi{https://doi.org/10.1007/s13280-016-0772-y}.
\bibitem[{Dal~Molin and Rizzoli(2022)}]{ricardo_cnn_s1_2022}
\bibinfo{author}{Dal~Molin, R.}, \bibinfo{author}{Rizzoli, P.}, \bibinfo{year}{2022}.
\newblock \bibinfo{title}{Potential of convolutional neural networks for forest mapping using {Sentinel-1} interferometric short time series}.
\newblock \bibinfo{journal}{Remote Sensing} \bibinfo{volume}{14}.
\newblock \URLprefix \url{https://www.mdpi.com/2072-4292/14/6/1381}, \DOIprefix\doi{10.3390/rs14061381}.
\bibitem[{Diniz et~al.(2013)Diniz, Kok, Hott, Hoogstra-Klein and Arts}]{amazonas_arc_defo2013}
\bibinfo{author}{Diniz, F.}, \bibinfo{author}{Kok, K.}, \bibinfo{author}{Hott, M.}, \bibinfo{author}{Hoogstra-Klein, M.}, \bibinfo{author}{Arts, B.}, \bibinfo{year}{2013}.
\newblock \bibinfo{title}{From space and from the ground: Determining forest dynamics in settlement projects in the {Brazilian Amazon}}.
\newblock \bibinfo{journal}{International Forestry Review} \bibinfo{volume}{15}, \bibinfo{pages}{442--455}.
\newblock \DOIprefix\doi{10.1505/146554813809025658}.
\bibitem[{Dos-Santos et~al.(2019)Dos-Santos, Keller and Morton}]{patches_amazonas}
\bibinfo{author}{Dos-Santos, M.}, \bibinfo{author}{Keller, M.}, \bibinfo{author}{Morton, D.}, \bibinfo{year}{2019}.
\newblock \bibinfo{title}{{LiDAR surveys over selected forest research sites, Brazilian Amazon, 2008-2018}}.
\newblock \URLprefix \url{https://daac.ornl.gov/cgi-bin/dsviewer.pl?ds_id=1644}, \DOIprefix\doi{10.3334/ORNLDAAC/1644}.
\bibitem[{FAO(2020)}]{fao_global_2020}
\bibinfo{author}{FAO}, \bibinfo{year}{2020}.
\newblock \bibinfo{title}{Global Forest Resources Assessment 2020}.
\newblock \bibinfo{publisher}{FAO}.
\newblock \URLprefix \url{http://www.fao.org/documents/card/en/c/ca9825en}, \DOIprefix\doi{10.4060/ca9825en}.
\bibitem[{Fassnacht et~al.(2023)Fassnacht, White, Wulder and Næsset}]{10.1093/forestry/cpad024}
\bibinfo{author}{Fassnacht, F.}, \bibinfo{author}{White, J.}, \bibinfo{author}{Wulder, M.}, \bibinfo{author}{Næsset, E.}, \bibinfo{year}{2023}.
\newblock \bibinfo{title}{{Remote sensing in forestry: current challenges, considerations and directions}}.
\newblock \bibinfo{journal}{Forestry: An International Journal of Forest Research} \bibinfo{volume}{97}, \bibinfo{pages}{11--37}.
\newblock \DOIprefix\doi{10.1093/forestry/cpad024}.
\bibitem[{Fritz et~al.(2012)Fritz, Breit, Rossi, Balss, Lachaise and Duque}]{fritz_itp_2012}
\bibinfo{author}{Fritz, T.}, \bibinfo{author}{Breit, H.}, \bibinfo{author}{Rossi, C.}, \bibinfo{author}{Balss, U.}, \bibinfo{author}{Lachaise, M.}, \bibinfo{author}{Duque, S.}, \bibinfo{year}{2012}.
\newblock \bibinfo{title}{Interferometric processing and products of the {TanDEM}-{X} mission}, in: \bibinfo{booktitle}{2012 {IEEE} {International} {Geoscience} and {Remote} {Sensing} {Symposium}}, \bibinfo{publisher}{IEEE}, \bibinfo{address}{Munich, Germany}. pp. \bibinfo{pages}{1904--1907}.
\newblock \URLprefix \url{http://ieeexplore.ieee.org/document/6351133/}, \DOIprefix\doi{10.1109/IGARSS.2012.6351133}.
\bibitem[{Gawlikowski et~al.(2022)Gawlikowski, Ebel, Schmitt and Zhu}]{gawlikowski_clouds_2022}
\bibinfo{author}{Gawlikowski, J.}, \bibinfo{author}{Ebel, P.}, \bibinfo{author}{Schmitt, M.}, \bibinfo{author}{Zhu, X.X.}, \bibinfo{year}{2022}.
\newblock \bibinfo{title}{Explaining the effects of clouds on remote sensing scene classification}.
\newblock \bibinfo{journal}{IEEE Journal of Selected Topics in Applied Earth Observations and Remote Sensing} \bibinfo{volume}{15}, \bibinfo{pages}{9976--9986}.
\newblock \URLprefix \url{https://ieeexplore.ieee.org/document/9956865/}, \DOIprefix\doi{10.1109/JSTARS.2022.3221788}.
\bibitem[{Gong et~al.(2019)Gong, Liu, Zhang, Li, Wang, Huang, Clinton, Ji, Li, Bai, Chen, Xu, Zhu, Yuan, {Ping Suen}, Guo, Xu, Li, Zhao, Yang, Yu, Wang, Fu, Yu, Dronova, Hui, Cheng, Shi, Xiao, Liu and Song}]{gong2019FROM-GLC}
\bibinfo{author}{Gong, P.}, \bibinfo{author}{Liu, H.}, \bibinfo{author}{Zhang, M.}, \bibinfo{author}{Li, C.}, \bibinfo{author}{Wang, J.}, \bibinfo{author}{Huang, H.}, \bibinfo{author}{Clinton, N.}, \bibinfo{author}{Ji, L.}, \bibinfo{author}{Li, W.}, \bibinfo{author}{Bai, Y.}, \bibinfo{author}{Chen, B.}, \bibinfo{author}{Xu, B.}, \bibinfo{author}{Zhu, Z.}, \bibinfo{author}{Yuan, C.}, \bibinfo{author}{{Ping Suen}, H.}, \bibinfo{author}{Guo, J.}, \bibinfo{author}{Xu, N.}, \bibinfo{author}{Li, W.}, \bibinfo{author}{Zhao, Y.}, \bibinfo{author}{Yang, J.}, \bibinfo{author}{Yu, C.}, \bibinfo{author}{Wang, X.}, \bibinfo{author}{Fu, H.}, \bibinfo{author}{Yu, L.}, \bibinfo{author}{Dronova, I.}, \bibinfo{author}{Hui, F.}, \bibinfo{author}{Cheng, X.}, \bibinfo{author}{Shi, X.}, \bibinfo{author}{Xiao, F.}, \bibinfo{author}{Liu, Q.}, \bibinfo{author}{Song, L.}, \bibinfo{year}{2019}.
\newblock \bibinfo{title}{Stable classification with limited sample: transferring a 30-m resolution sample set collected in 2015 to mapping 10-m resolution global land cover in 2017}.
\newblock \bibinfo{journal}{Science Bulletin} \bibinfo{volume}{64}, \bibinfo{pages}{370--373}.
\bibitem[{Gonzalez et~al.(2020)Gonzalez, Bachmann, Bueso-Bello, Rizzoli and Zink}]{gonzalez_editing_2020}
\bibinfo{author}{Gonzalez, C.}, \bibinfo{author}{Bachmann, M.}, \bibinfo{author}{Bueso-Bello, J.L.}, \bibinfo{author}{Rizzoli, P.}, \bibinfo{author}{Zink, M.}, \bibinfo{year}{2020}.
\newblock \bibinfo{title}{A fully automatic algorithm for editing the {TanDEM-X global DEM}}.
\newblock \bibinfo{journal}{Remote Sensing} \bibinfo{volume}{12}, \bibinfo{pages}{3961}.
\newblock \URLprefix \url{https://www.mdpi.com/2072-4292/12/23/3961}, \DOIprefix\doi{10.3390/rs12233961}.
\bibitem[{Gonzalez and Rizzoli(2018)}]{gonzalez2018}
\bibinfo{author}{Gonzalez, C.}, \bibinfo{author}{Rizzoli, P.}, \bibinfo{year}{2018}.
\newblock \bibinfo{title}{Landcover-dependent assessment of the relative height accuracy in {T}an{DEM-X} {DEM} products}.
\newblock \bibinfo{journal}{IEEE Geoscience Remote Sensing Letters} \bibinfo{volume}{15}, \bibinfo{pages}{1892--1896}.
\bibitem[{Hansen et~al.(2020)Hansen, Mitchard and King}]{hansen_s1_fnf_2020}
\bibinfo{author}{Hansen, J.}, \bibinfo{author}{Mitchard, E.}, \bibinfo{author}{King, S.}, \bibinfo{year}{2020}.
\newblock \bibinfo{title}{Assessing forest/non-forest separability using {Sentinel-1} {C}-band synthetic aperture radar}.
\newblock \bibinfo{journal}{Remote Sensing} \bibinfo{volume}{12}.
\newblock \URLprefix \url{https://www.mdpi.com/2072-4292/12/11/1899}, \DOIprefix\doi{10.3390/rs12111899}.
\bibitem[{Hansen et~al.(2013)Hansen, Potapov, Moore, Hancher, Turubanova, Tyukavina, Thau, Stehamn, Goetz, Loveland and Kommareddy}]{Hansen2013}
\bibinfo{author}{Hansen, M.C.}, \bibinfo{author}{Potapov, P.V.}, \bibinfo{author}{Moore, R.}, \bibinfo{author}{Hancher, M.}, \bibinfo{author}{Turubanova, S.A.}, \bibinfo{author}{Tyukavina, A.}, \bibinfo{author}{Thau, D.}, \bibinfo{author}{Stehamn, S.V.}, \bibinfo{author}{Goetz, S.J.}, \bibinfo{author}{Loveland, T.R.}, \bibinfo{author}{Kommareddy, J.R.G.}, \bibinfo{year}{2013}.
\newblock \bibinfo{title}{High-resolution global maps of 21st century forest coverage change}.
\newblock \bibinfo{journal}{Science} \bibinfo{volume}{342}, \bibinfo{pages}{850--853}.
\bibitem[{Jadon(2020)}]{Jadon2020loss_functions_semantic_segmentation}
\bibinfo{author}{Jadon, S.}, \bibinfo{year}{2020}.
\newblock \bibinfo{title}{A survey of loss functions for semantic segmentation}.
\newblock \bibinfo{journal}{{IEEE Conference on Computational Intelligence in Bioinformatics and Computational Biology (CIBCB)}} , \bibinfo{pages}{1--7}.
\bibitem[{Krieger et~al.(2007)Krieger, Moreira, Fiedler, Hajnsek, Werner, Younis and Zink}]{krieger2007tandem}
\bibinfo{author}{Krieger, G.}, \bibinfo{author}{Moreira, A.}, \bibinfo{author}{Fiedler, H.}, \bibinfo{author}{Hajnsek, I.}, \bibinfo{author}{Werner, M.}, \bibinfo{author}{Younis, M.}, \bibinfo{author}{Zink, M.}, \bibinfo{year}{2007}.
\newblock \bibinfo{title}{Tan{DEM-X}: A satellite formation for high-resolution {SAR} interferometry}.
\newblock \bibinfo{journal}{IEEE Transactions on Geoscience and Remote Sensing} \bibinfo{volume}{45}.
\bibitem[{Ma et~al.(2019)Ma, Liu, Zhang, Ye, Yin and Johnson}]{ma2019DLRS}
\bibinfo{author}{Ma, L.}, \bibinfo{author}{Liu, Y.}, \bibinfo{author}{Zhang, X.}, \bibinfo{author}{Ye, Y.}, \bibinfo{author}{Yin, G.}, \bibinfo{author}{Johnson, B.}, \bibinfo{year}{2019}.
\newblock \bibinfo{title}{{Deep learning in remote sensing applications: A meta-analysis and review}}.
\newblock \bibinfo{journal}{ISPRS Journal of Photogrammetry and Remote Sensing} \bibinfo{volume}{152}, \bibinfo{pages}{166--177}.
\bibitem[{Martone et~al.(2012)Martone, Br\"{a}utigam, Rizzoli, Gonzalez, Bachmann and Krieger}]{martone2012Coh_ISPRS}
\bibinfo{author}{Martone, M.}, \bibinfo{author}{Br\"{a}utigam, B.}, \bibinfo{author}{Rizzoli, P.}, \bibinfo{author}{Gonzalez, C.}, \bibinfo{author}{Bachmann, M.}, \bibinfo{author}{Krieger, G.}, \bibinfo{year}{2012}.
\newblock \bibinfo{title}{{Coherence evaluation of TanDEM-X interferometric data}}.
\newblock \bibinfo{journal}{ISPRS Journal of Photogrammetry and Remote Sensing} \bibinfo{volume}{73}, \bibinfo{pages}{21--29}.
\bibitem[{Martone et~al.(2016)Martone, Rizzoli and Krieger}]{martone2016volume}
\bibinfo{author}{Martone, M.}, \bibinfo{author}{Rizzoli, P.}, \bibinfo{author}{Krieger, G.}, \bibinfo{year}{2016}.
\newblock \bibinfo{title}{Volume decorrelation effects in {T}an{DEM-X} interferometric {SAR} data}.
\newblock \bibinfo{journal}{IEEE Geoscience and Remote Sensing Letters} \bibinfo{volume}{13}, \bibinfo{pages}{1812--1816}.
\bibitem[{Martone et~al.(2018a)Martone, Rizzoli, Wecklich, Gonzalez, Bueso-Bello, Valdo, Schulze, Zink, Krieger and Moreira}]{martone2018FNF}
\bibinfo{author}{Martone, M.}, \bibinfo{author}{Rizzoli, P.}, \bibinfo{author}{Wecklich, C.}, \bibinfo{author}{Gonzalez, C.}, \bibinfo{author}{Bueso-Bello, J.L.}, \bibinfo{author}{Valdo, P.}, \bibinfo{author}{Schulze, D.}, \bibinfo{author}{Zink, M.}, \bibinfo{author}{Krieger, G.}, \bibinfo{author}{Moreira, A.}, \bibinfo{year}{2018}a.
\newblock \bibinfo{title}{{The global forest/non-forest map from TanDEM-X interferometric SAR data}}.
\newblock \bibinfo{journal}{Remote Sensing of Environment} \bibinfo{volume}{205}, \bibinfo{pages}{352--373}.
\bibitem[{Martone et~al.(2018b)Martone, Sica, Gonzalez, Bueso-Bello, Valdo and Rizzoli}]{martone2018FNFhighres}
\bibinfo{author}{Martone, M.}, \bibinfo{author}{Sica, F.}, \bibinfo{author}{Gonzalez, C.}, \bibinfo{author}{Bueso-Bello, J.L.}, \bibinfo{author}{Valdo, P.}, \bibinfo{author}{Rizzoli, P.}, \bibinfo{year}{2018}b.
\newblock \bibinfo{title}{{High-resolution forest mapping from TanDEM-X interferometric data exploiting nonlocal filtering}}.
\newblock \bibinfo{journal}{Remote Sensing} \bibinfo{volume}{10}, \bibinfo{pages}{1477}.
\bibitem[{Mazza et~al.(2019)Mazza, Sica, Rizzoli and Scarpa}]{Mazza2019}
\bibinfo{author}{Mazza, A.}, \bibinfo{author}{Sica, F.}, \bibinfo{author}{Rizzoli, P.}, \bibinfo{author}{Scarpa, G.}, \bibinfo{year}{2019}.
\newblock \bibinfo{title}{{TanDEM-X} forest mapping using convolutional neural networks}.
\newblock \bibinfo{journal}{Remote Sensing} \bibinfo{volume}{11}.
\bibitem[{O’Neil-Dunne et~al.(2014)O’Neil-Dunne, MacFaden, Royar, Reis, Dubayah and Swatantran}]{lidar_data}
\bibinfo{author}{O’Neil-Dunne, J.}, \bibinfo{author}{MacFaden, S.}, \bibinfo{author}{Royar, A.}, \bibinfo{author}{Reis, M.}, \bibinfo{author}{Dubayah, R.}, \bibinfo{author}{Swatantran, A.}, \bibinfo{year}{2014}.
\newblock \bibinfo{title}{An object-based approach to statewide land cover mapping}, in: \bibinfo{booktitle}{Proceedings of ASPRS 2014 annual conference}, pp. \bibinfo{pages}{23--28}.
\bibitem[{Paszke et~al.(2019)Paszke, Gross, Massa, Lerer, Bradbury, Chanan, Killeen, Lin, Gimelshein, Antiga, Desmaison, K{\"{o}}pf, Yang, DeVito, Raison, Tejani, Chilamkurthy, Steiner, Fang, Bai and Chintala}]{torch}
\bibinfo{author}{Paszke, A.}, \bibinfo{author}{Gross, S.}, \bibinfo{author}{Massa, F.}, \bibinfo{author}{Lerer, A.}, \bibinfo{author}{Bradbury, J.}, \bibinfo{author}{Chanan, G.}, \bibinfo{author}{Killeen, T.}, \bibinfo{author}{Lin, Z.}, \bibinfo{author}{Gimelshein, N.}, \bibinfo{author}{Antiga, L.}, \bibinfo{author}{Desmaison, A.}, \bibinfo{author}{K{\"{o}}pf, A.}, \bibinfo{author}{Yang, E.}, \bibinfo{author}{DeVito, Z.}, \bibinfo{author}{Raison, M.}, \bibinfo{author}{Tejani, A.}, \bibinfo{author}{Chilamkurthy, S.}, \bibinfo{author}{Steiner, B.}, \bibinfo{author}{Fang, L.}, \bibinfo{author}{Bai, J.}, \bibinfo{author}{Chintala, S.}, \bibinfo{year}{2019}.
\newblock \bibinfo{title}{{PyTorch: An} Imperative Style, High-Performance Deep Learning Library}. \bibinfo{publisher}{Curran Associates Inc.}, \bibinfo{address}{Red Hook, NY, USA}.
\bibitem[{Pathak et~al.(2016)Pathak, Kr{\"{a}}henb{\"{u}}hl, Donahue, Darrell and Efros}]{pathak2016context_encoders_inpainting}
\bibinfo{author}{Pathak, D.}, \bibinfo{author}{Kr{\"{a}}henb{\"{u}}hl, P.}, \bibinfo{author}{Donahue, J.}, \bibinfo{author}{Darrell, T.}, \bibinfo{author}{Efros, A.}, \bibinfo{year}{2016}.
\newblock \bibinfo{title}{Context encoders: Feature learning by inpainting}.
\newblock \bibinfo{journal}{CoRR} \URLprefix \url{http://arxiv.org/abs/1604.07379}.
\bibitem[{Pulella et~al.(2024)Pulella, Prats-Iraola and Sica}]{Pulella2024}
\bibinfo{author}{Pulella, A.}, \bibinfo{author}{Prats-Iraola, P.}, \bibinfo{author}{Sica, F.}, \bibinfo{year}{2024}.
\newblock \bibinfo{title}{Multitask learning for phase source separation in {InSAR} burst modes}.
\newblock \bibinfo{journal}{IEEE Transactions on Geoscience and Remote Sensing} \bibinfo{volume}{62}, \bibinfo{pages}{1--21}.
\newblock \DOIprefix\doi{10.1109/TGRS.2024.3401775}.
\bibitem[{Rizzoli et~al.(2022)Rizzoli, Dell'Amore, Bueso-Bello, Gollin, Carcereri and Martone}]{rizzoli_volume_2022}
\bibinfo{author}{Rizzoli, P.}, \bibinfo{author}{Dell'Amore, L.}, \bibinfo{author}{Bueso-Bello, J.L.}, \bibinfo{author}{Gollin, N.}, \bibinfo{author}{Carcereri, D.}, \bibinfo{author}{Martone, M.}, \bibinfo{year}{2022}.
\newblock \bibinfo{title}{{On the derivation of volume decorrelation from TanDEM-X bistatic coherence}}.
\newblock \bibinfo{journal}{IEEE Journal of Selected Topics in Applied Earth Observations and Remote Sensing} \bibinfo{volume}{15}, \bibinfo{pages}{3504--3518}.
\newblock \URLprefix \url{https://ieeexplore.ieee.org/document/9763376/}, \DOIprefix\doi{10.1109/JSTARS.2022.3170076}.
\bibitem[{Rizzoli et~al.(2017)Rizzoli, Martone, Gonzalez, Wecklich, Br\"{a}utigam, Borla~Tridon, Bachmann, Schulze, Fritz, Huber, Wessel, Krieger, Zink and Moreira}]{rizzoli2016DEMperformance}
\bibinfo{author}{Rizzoli, P.}, \bibinfo{author}{Martone, M.}, \bibinfo{author}{Gonzalez, C.}, \bibinfo{author}{Wecklich, C.}, \bibinfo{author}{Br\"{a}utigam, B.}, \bibinfo{author}{Borla~Tridon, D.}, \bibinfo{author}{Bachmann, M.}, \bibinfo{author}{Schulze, D.}, \bibinfo{author}{Fritz, T.}, \bibinfo{author}{Huber, M.}, \bibinfo{author}{Wessel, B.}, \bibinfo{author}{Krieger, G.}, \bibinfo{author}{Zink, M.}, \bibinfo{author}{Moreira, A.}, \bibinfo{year}{2017}.
\newblock \bibinfo{title}{{Generation and performance assessment of the global TanDEM-X digital elevation model}}.
\newblock \bibinfo{journal}{ISPRS Journal of Photogrammetry and Remote Sensing} \bibinfo{volume}{132}, \bibinfo{pages}{119--139}.
\bibitem[{Ronneberger et~al.(2015)Ronneberger, Fischer and Brox}]{ronneberger2015unet}
\bibinfo{author}{Ronneberger, O.}, \bibinfo{author}{Fischer, P.}, \bibinfo{author}{Brox, T.}, \bibinfo{year}{2015}.
\newblock \bibinfo{title}{{U-Net}: Convolutional networks for biomedical image segmentation}.
\bibitem[{Schepaschenko et~al.(2021)Schepaschenko, Moltchanova, Fedorov, Karminov, Ontikov, Santoro, See, Kositsyn, Shvidenko, Romanovskaya, Korotkov, Lesiv, Bartalev, Fritz, Shchepashchenko and Kraxner}]{Schepaschenko2021}
\bibinfo{author}{Schepaschenko, D.}, \bibinfo{author}{Moltchanova, E.}, \bibinfo{author}{Fedorov, S.}, \bibinfo{author}{Karminov, V.}, \bibinfo{author}{Ontikov, P.}, \bibinfo{author}{Santoro, M.}, \bibinfo{author}{See, L.}, \bibinfo{author}{Kositsyn, V.}, \bibinfo{author}{Shvidenko, A.}, \bibinfo{author}{Romanovskaya, A.}, \bibinfo{author}{Korotkov, V.}, \bibinfo{author}{Lesiv, M.}, \bibinfo{author}{Bartalev, S.}, \bibinfo{author}{Fritz, S.}, \bibinfo{author}{Shchepashchenko, M.}, \bibinfo{author}{Kraxner, F.}, \bibinfo{year}{2021}.
\newblock \bibinfo{title}{Russian forest sequesters substantially more carbon than previously reported}.
\newblock \bibinfo{journal}{Scientific Reports} \bibinfo{volume}{11}, \bibinfo{pages}{12825}.
\newblock \DOIprefix\doi{10.1038/s41598-021-92152-9}.
\bibitem[{Schlund et~al.(2014)Schlund, von Poncet, Hoekman, Kuntz and Schmullius}]{SCHLUND201416}
\bibinfo{author}{Schlund, M.}, \bibinfo{author}{von Poncet, F.}, \bibinfo{author}{Hoekman, D.}, \bibinfo{author}{Kuntz, S.}, \bibinfo{author}{Schmullius, C.}, \bibinfo{year}{2014}.
\newblock \bibinfo{title}{{Importance of bistatic SAR features from TanDEM-X for forest mapping and monitoring}}.
\newblock \bibinfo{journal}{Remote Sensing of Environment} \bibinfo{volume}{151}, \bibinfo{pages}{16--26}.
\newblock \bibinfo{note}{Special Issue on 2012 ForestSAT}.
\bibitem[{Shimada et~al.(2014)Shimada, Itoh, Motooka, Watanabe, Shiraishi, Thapa and Lucas}]{shimada2014forest}
\bibinfo{author}{Shimada, M.}, \bibinfo{author}{Itoh, T.}, \bibinfo{author}{Motooka, T.}, \bibinfo{author}{Watanabe, M.}, \bibinfo{author}{Shiraishi, T.}, \bibinfo{author}{Thapa, R.}, \bibinfo{author}{Lucas, R.}, \bibinfo{year}{2014}.
\newblock \bibinfo{title}{New global forest/non-forest maps from {ALOS} {PALSAR} data (2007-2010)}.
\newblock \bibinfo{journal}{Remote Sensing of Environment} \bibinfo{volume}{155}, \bibinfo{pages}{13--31}.
\bibitem[{Sica et~al.(2022)Sica, Calvanese, Scarpa and Rizzoli}]{Sica2022}
\bibinfo{author}{Sica, F.}, \bibinfo{author}{Calvanese, F.}, \bibinfo{author}{Scarpa, G.}, \bibinfo{author}{Rizzoli, P.}, \bibinfo{year}{2022}.
\newblock \bibinfo{title}{A coherence-driven approach for {I}n{SAR} phase unwrapping using convolutional neural network}.
\newblock \bibinfo{journal}{IEEE Geoscience and Remote Sensing Letters} \bibinfo{volume}{19}, \bibinfo{pages}{1--5}.
\bibitem[{Sica et~al.(2020)Sica, Gobbi, Rizzoli and Bruzzone}]{phinet}
\bibinfo{author}{Sica, F.}, \bibinfo{author}{Gobbi, G.}, \bibinfo{author}{Rizzoli, P.}, \bibinfo{author}{Bruzzone, L.}, \bibinfo{year}{2020}.
\newblock \bibinfo{title}{$\phi$-net: Deep residual learning for {InSAR} parameters estimation}.
\newblock \bibinfo{journal}{IEEE Transactions on Geoscience and Remote Sensing} \bibinfo{volume}{PP}.
\newblock \DOIprefix\doi{10.1109/TGRS.2020.3020427}.
\bibitem[{Singh et~al.(2018)Singh, Batra, Pang, Torresani, Basu, Paluri and Jawahar}]{singh2018ssl_semantic_segmentation_overhead_imagery}
\bibinfo{author}{Singh, S.}, \bibinfo{author}{Batra, A.}, \bibinfo{author}{Pang, G.}, \bibinfo{author}{Torresani, L.}, \bibinfo{author}{Basu, S.}, \bibinfo{author}{Paluri, M.}, \bibinfo{author}{Jawahar, C.}, \bibinfo{year}{2018}.
\newblock \bibinfo{title}{Self-supervised feature learning for semantic segmentation of overhead imagery}, in: \bibinfo{booktitle}{British Machine Vision Conference}, pp. \bibinfo{pages}{1--13}.
\bibitem[{Sudre et~al.(2017)Sudre, Li, Vercauteren, Ourselin and Cardoso}]{dice}
\bibinfo{author}{Sudre, C.}, \bibinfo{author}{Li, W.}, \bibinfo{author}{Vercauteren, T.}, \bibinfo{author}{Ourselin, S.}, \bibinfo{author}{Cardoso, M.}, \bibinfo{year}{2017}.
\newblock \bibinfo{title}{Generalised dice overlap as a deep learning loss function for highly unbalanced segmentations}, in: \bibinfo{booktitle}{Deep Learning in Medical Image Analysis and Multimodal Learning for Clinical Decision Support}, \bibinfo{publisher}{Springer International Publishing}, \bibinfo{address}{Cham}. pp. \bibinfo{pages}{240--248}.
\bibitem[{Townsend(1971)}]{townsend1971theoretical}
\bibinfo{author}{Townsend, J.}, \bibinfo{year}{1971}.
\newblock \bibinfo{title}{Theoretical analysis of an alphabetic confusion matrix}.
\newblock \bibinfo{journal}{Perception \& Psychophysics} \bibinfo{volume}{9}, \bibinfo{pages}{40--50}.
\bibitem[{UNFCCC(2020)}]{unfccc_2020}
\bibinfo{author}{UNFCCC}, \bibinfo{year}{2020}.
\newblock \bibinfo{title}{United Nations Climate Change, Annual Report 2020}.
\newblock \bibinfo{publisher}{United Nations Framework Convention on Climate Change}.
\bibitem[{Wang et~al.(2022)Wang, Albrecht, Braham, Mou and Zhu}]{ssl_survey}
\bibinfo{author}{Wang, Y.}, \bibinfo{author}{Albrecht, C.}, \bibinfo{author}{Braham, N.}, \bibinfo{author}{Mou, L.}, \bibinfo{author}{Zhu, X.}, \bibinfo{year}{2022}.
\newblock \bibinfo{title}{Self-supervised learning in remote sensing: A review}.
\newblock \href{http://arxiv.org/abs/2206.13188}{{\tt arXiv:2206.13188}}.
\bibitem[{Zanaga et~al.(2022)Zanaga, Van De~Kerchove, Daems, De~Keersmaecker, Brockmann, Kirches, Wevers, Cartus, Santoro, Fritz, Lesiv, Herold, Tsendbazar, Xu, Ramoino and Arino}]{esa_worldcover_2021}
\bibinfo{author}{Zanaga, D.}, \bibinfo{author}{Van De~Kerchove, R.}, \bibinfo{author}{Daems, D.}, \bibinfo{author}{De~Keersmaecker, W.}, \bibinfo{author}{Brockmann, C.}, \bibinfo{author}{Kirches, G.}, \bibinfo{author}{Wevers, J.}, \bibinfo{author}{Cartus, O.}, \bibinfo{author}{Santoro, M.}, \bibinfo{author}{Fritz, S.}, \bibinfo{author}{Lesiv, M.}, \bibinfo{author}{Herold, M.}, \bibinfo{author}{Tsendbazar, N.}, \bibinfo{author}{Xu, P.}, \bibinfo{author}{Ramoino, F.}, \bibinfo{author}{Arino, O.}, \bibinfo{year}{2022}.
\newblock \bibinfo{title}{{ESA WorldCover 10 m 2021 v200}}.
\newblock \bibinfo{journal}{Zenodo} .
\bibitem[{Zanaga et~al.(2021)Zanaga, Van De~Kerchove, De~Keersmaecker, Souverijns, Brockmann, Quast, Wevers, Grosu, Paccini, Vergnaud, Cartus, Santoro, Fritz, Georgieva, Lesiv, Carter, Herold, Li, Tsendbazar, Ramoino and Arino}]{esa_worldcover_2020}
\bibinfo{author}{Zanaga, D.}, \bibinfo{author}{Van De~Kerchove, R.}, \bibinfo{author}{De~Keersmaecker, W.}, \bibinfo{author}{Souverijns, N.}, \bibinfo{author}{Brockmann, C.}, \bibinfo{author}{Quast, R.}, \bibinfo{author}{Wevers, J.}, \bibinfo{author}{Grosu, A.}, \bibinfo{author}{Paccini, A.}, \bibinfo{author}{Vergnaud, S.}, \bibinfo{author}{Cartus, O.}, \bibinfo{author}{Santoro, M.}, \bibinfo{author}{Fritz, S.}, \bibinfo{author}{Georgieva, I.}, \bibinfo{author}{Lesiv, M.}, \bibinfo{author}{Carter, S.}, \bibinfo{author}{Herold, M.}, \bibinfo{author}{Li, L.}, \bibinfo{author}{Tsendbazar, N.E.}, \bibinfo{author}{Ramoino, F.}, \bibinfo{author}{Arino, O.}, \bibinfo{year}{2021}.
\newblock \bibinfo{title}{{ESA WorldCover 10 m 2020 v100}}.
\newblock \bibinfo{journal}{Zenodo} .
\bibitem[{Zebker and Villasenor(1992)}]{zebker1992decorrelation}
\bibinfo{author}{Zebker, H.}, \bibinfo{author}{Villasenor, J.}, \bibinfo{year}{1992}.
\newblock \bibinfo{title}{{Decorrelation in interferometric radar echoes}}.
\newblock \bibinfo{journal}{IEEE Transactions on Geoscience and Remote Sensing} \bibinfo{volume}{30}, \bibinfo{pages}{950--959}.
\bibitem[{Zhu et~al.(2021)Zhu, Montazeri, Ali, Hua, Wang, Mou, Shi, Xu and Bamler}]{zhu2021DLSAR}
\bibinfo{author}{Zhu, X.}, \bibinfo{author}{Montazeri, S.}, \bibinfo{author}{Ali, M.}, \bibinfo{author}{Hua, Y.}, \bibinfo{author}{Wang, Y.}, \bibinfo{author}{Mou, L.}, \bibinfo{author}{Shi, Y.}, \bibinfo{author}{Xu, F.}, \bibinfo{author}{Bamler, R.}, \bibinfo{year}{2021}.
\newblock \bibinfo{title}{Deep learning meets {SAR}: Concepts, models, pitfalls, and perspectives}.
\newblock \bibinfo{journal}{IEEE Geoscience and Remote Sensing Magazine} \bibinfo{volume}{9}, \bibinfo{pages}{143--172}.
\bibitem[{Zhu et~al.(2017)Zhu, Tuia, Mou, Xia, Zhang, Xu and Fraundorfer}]{zhu2017DLinSAR}
\bibinfo{author}{Zhu, X.}, \bibinfo{author}{Tuia, D.}, \bibinfo{author}{Mou, L.}, \bibinfo{author}{Xia, G.}, \bibinfo{author}{Zhang, L.}, \bibinfo{author}{Xu, F.}, \bibinfo{author}{Fraundorfer, F.}, \bibinfo{year}{2017}.
\newblock \bibinfo{title}{Deep learning in remote sensing: A comprehensive review and list of resources}.
\newblock \bibinfo{journal}{IEEE Geoscience and Remote Sensing Magazine} \bibinfo{volume}{5}, \bibinfo{pages}{8--36}.
\bibitem[{Zink et~al.(2021)Zink, Moreira, Hajnsek, Rizzoli, Bachmann, Kahle, Fritz, Huber, Krieger, Lachaise, Martone, Maurer and Wessel}]{zink_tdm_2021}
\bibinfo{author}{Zink, M.}, \bibinfo{author}{Moreira, A.}, \bibinfo{author}{Hajnsek, I.}, \bibinfo{author}{Rizzoli, P.}, \bibinfo{author}{Bachmann, M.}, \bibinfo{author}{Kahle, R.}, \bibinfo{author}{Fritz, T.}, \bibinfo{author}{Huber, M.}, \bibinfo{author}{Krieger, G.}, \bibinfo{author}{Lachaise, M.}, \bibinfo{author}{Martone, M.}, \bibinfo{author}{Maurer, E.}, \bibinfo{author}{Wessel, B.}, \bibinfo{year}{2021}.
\newblock \bibinfo{title}{Tan{DEM-X}: 10 years of formation flying bistatic {SAR} interferometry}.
\newblock \bibinfo{journal}{IEEE Journal of Selected Topics in Applied Earth Observations and Remote Sensing} \bibinfo{volume}{14}, \bibinfo{pages}{3546--3565}.

\end{thebibliography}
\end{document}